\documentclass[]{fairmeta}

\usepackage{amssymb}
\usepackage{pifont}
\usepackage{multicol}
\usepackage{enumitem}
\usepackage{makecell}
\usepackage{multirow}
\usepackage{subcaption}

\title{GISTBench: Evaluating LLM User Understanding via Evidence-Based Interest Verification}

\author[1,*,\dagger]{Iordanis Fostiropoulos}
\author[1]{Muhammad Rafay Azhar}
\author[1]{Abdalaziz Sawwan}
\author[1]{Boyu Fang}
\author[1]{Yuchen Liu}
\author[1]{Jiayi Liu}
\author[1,\dagger]{Hanchao Yu}
\author[1,\dagger]{Qi Guo}
\author[1]{Jianyu Wang}
\author[1]{Fei Liu}
\author[1,*]{Xiangjun Fan}

\affiliation[1]{Meta Recommendation Systems (MRS)}

\contribution[*]{Corresponding author}
\contribution[\dagger]{Project leads}

\abstract{We introduce GISTBench, a benchmark for evaluating Large Language Models' (LLMs) ability to understand users from their interaction histories in recommendation systems. Unlike traditional RecSys benchmarks that focus on item prediction accuracy, our benchmark evaluates how well LLMs can extract and verify user interests from engagement data. We propose two novel metric families: Interest Groundedness (IG), decomposed into precision and recall components to separately penalize hallucinated interest categories and reward coverage, and Interest Specificity (IS), which assesses the distinctiveness of verified LLM-predicted user profiles. We release a synthetic dataset constructed on real user interactions on a global short-form video platform. Our dataset contains both implicit and explicit engagement signals and rich textual descriptions. We validate our dataset fidelity against user surveys, and evaluate eight open-weight LLMs spanning 7B to 235B parameters, together with three proprietary frontier models (GPT-5, Claude 4.6, and Gemini 3.5 Flash). Our findings reveal performance bottlenecks in current LLMs, particularly their limited ability to accurately count and attribute engagement signals across heterogeneous interaction types.}

\date{March 25, 2026}
\correspondence{Iordanis Fostiropoulos at \email{iordanis@meta.com}, Xiangjun Fan at \email{maxfan@meta.com}}

\metadata[Code]{\url{https://github.com/facebookresearch/GISTBench}}

\begin{document}

\maketitle

\section{Introduction}
\label{sec:intro}

The accurate and effective modeling of user preferences and profiles is a foundational challenge in modern personalized systems, particularly within large-scale recommendation systems (RecSys). This process, termed User Understanding, aims to construct a rich representation of user interests that serves multiple downstream applications beyond prediction. Generating this profile in a natural language modality is useful for several reasons: it allows integration into personalized LLM-driven experiences, it improves data-driven feedback loops for RecSys, it enables cross-platform user profile portability for conversational agents and other products~\cite{li2024personalization}, and it provides grounding context for conversational recommender systems and generative recommendation pipelines that require interpretable user models for multi-turn reasoning.

Despite rapid growth in LLM benchmarks for reasoning, coding, and knowledge retrieval, none evaluate whether LLMs can accurately extract user interests from behavioral interaction histories. Evaluating such text-based interest representations requires new metrics.
Existing approaches either optimize the profile for downstream recommendation accuracy or treat it as a static display artifact, without verifying whether the generated text is factually grounded in the user's behavioral history. Our benchmark does not evaluate whether predicted interests improve recommendations (a separate problem). Instead, it tests whether the model can correctly reason about engagement signals. A model that cannot count engagements, weight implicit versus explicit signals, or parse nuanced content metadata will fail our verification criteria regardless of how plausible its output reads. Traditional RecSys evaluation typically measures single-item engagement prediction (e.g., click-through rate, next-item ranking). Such evaluations conflate user interest with confounding factors like positional bias in ranked lists, temporal context such as time of day, and variation in content production quality. Our benchmark avoids these confounds by evaluating interest extraction directly: given a complete interaction history with heterogeneous engagement signals, can the model identify what the user genuinely cares about, independent of how individual items were surfaced?

Current evaluation paradigms do not address this task well. Existing xAI metrics, such as faithfulness and plausibility, are designed to evaluate a model's reasoning process rather than the precision and consistency of inferred user interests \cite{dwivedi2023explainable}. Establishing reliable ground-truth for User Understanding is difficult because of inter-annotator disagreement, the gap between a user's expressed and behavioral preferences, and the subjective nature of interest labeling \cite{swamy2024interpretcc}. The task also requires simultaneous processing of a user's entire interaction history (UIH) to ensure that extracted interests are sufficiently grounded in multiple pieces of evidence.

To address these limitations, this work introduces GISTBench (\textbf{G}rounded \textbf{I}nterest \textbf{S}pecificity with \textbf{T}axonomy-\textbf{B}alanced \textbf{E}valuation), a framework with the following contributions:
\begin{itemize}
    \item \textbf{A Novel User Understanding Dataset:} We release a synthetic dataset that captures a broad spectrum of user engagement via both implicit (e.g., watch time) and explicit (e.g., likes/dislikes) positive and negative signals. Our dataset is grounded on real user engagements on a global short-form video platform.
    \item \textbf{A Multi-Faceted Benchmark:} We establish a rigorous benchmark evaluating contemporary models on existing RecSys datasets (KuaiRec, MIND, Amazon Music, Goodreads) as well as our novel synthetic dataset, validated through user survey data.
    \item \textbf{Novel Verification Metrics:} We propose Interest Groundedness (IG), decomposed into precision and recall components via an ensemble oracle, and Interest Specificity (IS), computed over verified interest categories only. These metrics verify LLM-predicted user profiles without requiring ground-truth labels. They correlate with user surveys ($\rho = 0.67$) and expose two complementary failure modes: hallucination and incomplete coverage. Predicted interests are checked against observable behavioral signals rather than subjective labels or model self-assessment.
\end{itemize}

Our findings show that current LLMs struggle to accurately count and attribute engagement signals across heterogeneous interaction types, and also have difficulty with strict instruction following.

\begin{figure}[t]
    \centering
    \includegraphics[width=\columnwidth]{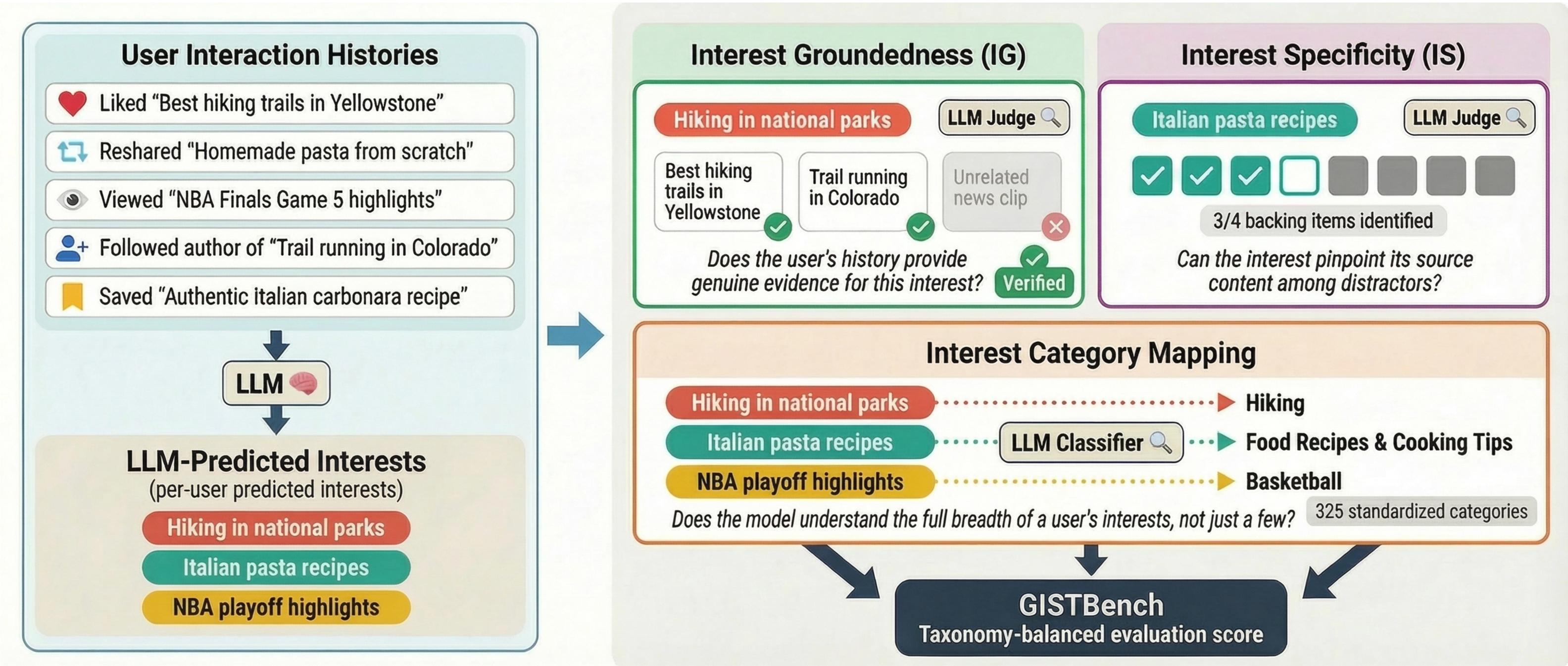}
    \caption{Overview of the GISTBench evaluation pipeline. User interaction histories are processed by an LLM to produce predicted interests. These are evaluated on two axes: Interest Groundedness (IG, decomposed into precision and recall) and Interest Specificity (IS, computed over verified categories only), then mapped to standardized taxonomy categories for normalization.}
    \label{fig:pipeline}
\end{figure}

\section{Related Work}
\label{sec:related}

\paragraph{LLMs for User Profiling and Representation.}
Recent literature has increasingly leveraged LLMs to transition user representation from latent embeddings to natural language. \citet{wei2024llmrec} introduce \emph{LLMRec}, which utilizes LLMs to generate synthetic interactions and denoise implicit feedback signals, thereby enriching the collaborative filtering graph to solve data sparsity. Similarly, \citet{gao2024end} propose \emph{LangPTune}, an end-to-end framework where LLMs are fine-tuned to generate user profiles that explicitly maximize downstream recommendation accuracy (e.g., Hit Rate). Moving towards transparency, \citet{ramos2024transparent} explore using LLMs to generate holistic, human-readable summaries of user preferences that serve as the primary interface for recommendation, allowing users to inspect and edit their own profiles.
These approaches treat the textual profile as an intermediate utility to improve item prediction or as a static display artifact. They do not verify whether the generated text is factually grounded in the user's history. Our benchmark instead evaluates the \emph{intrinsic quality} of the profile itself: its groundedness (IG) and distinctiveness (IS).

\paragraph{LLMs as Evaluators and Simulated Users.}
The concept of ``LLM-as-a-Judge'' has gained traction for evaluating subjective tasks. In the recommendation domain, \citet{wu2024recsys} introduce \emph{RecSys Arena}, where an LLM plays the role of a user to perform pairwise comparisons between different recommender outputs, mimicking human preference judgments. \citet{wang2023rethinking} extend this to conversational systems, using LLMs to simulate user interactions and evaluate the multi-turn dialogue flow rather than single-turn accuracy. General frameworks like \citet{zheng2023judging} have standardized the use of strong LLMs (e.g., GPT-4) to score the responses of weaker models, establishing best practices for prompt engineering to mitigate position bias.
These frameworks focus on evaluating the \emph{output item} (is this movie a good recommendation?) or the \emph{conversation quality}. They do not evaluate the \emph{reasoning process} of extracting user interests from raw logs. Standard LLM judges also suffer from hallucination. We address this by constraining the evaluator with strict evidence-counting tasks (IG), moving from open-ended ``judging'' to verifiable ``auditing.''

\paragraph{Explainability, Faithfulness, and Plausibility.}
Explainable Recommendation (xRec) aims to justify why an item was suggested. \citet{zhang2024large} demonstrate that GPT-4 can effectively evaluate the quality of natural language explanations, correlating well with human perception. Research in this area often navigates the trade-off between \emph{plausibility} (how convincing an explanation sounds to a human) and \emph{faithfulness} (how accurately it reflects the model's internal decision process \citep{jacovi2020towards, agarwal2024faithfulness}). \citet{lubos2024llm} and \citet{silva2024leveraging} conduct user studies showing that while LLM-predicted explanations are perceived as highly plausible and useful, users often cannot discern if the explanation is hallucinated.
However, existing metrics for faithfulness are typically tied to feature-importance weights in neural networks or rely on subjective user surveys which capture perceived plausibility rather than truth. No existing metric verifies textual explanations against behavioral data. We introduce IG to fill this role: it checks whether a predicted interest is supported by specific engagement signals in the User Interaction History (UIH), separating plausible hallucinations from grounded inferences.

\paragraph{Concurrent Work: Attribution-Level Benchmarks.}
Concurrently, \citet{ren2026alpbench} introduce \emph{ALPBench}, a benchmark for attribution-level long-term personal behavior understanding that evaluates LLMs on user behavioral data from a short-video platform using multi-dimensional labels and long-context evaluation. Our work differs in three respects. First, GISTBench evaluates \emph{interest groundedness} rather than attribution accuracy, verifying whether predicted interests are supported by sufficient engagement evidence through configurable signal thresholds without requiring ground-truth attribution labels. Second, our precision-recall decomposition of IG and the complementary IS metric separate hallucination from incomplete coverage and evaluate the discriminative specificity of verified interests. Third, GISTBench evaluates across five heterogeneous datasets spanning different domains and signal types, testing generalization across recommendation contexts rather than focusing on a single platform. The two benchmarks are complementary: ALPBench can diagnose whether a model correctly links a behavior to its source item, while GISTBench can diagnose whether the model accumulates enough such links to justify an interest claim.

\paragraph{LLM Agents and Synthetic Personas for Recommendation.}
Recent work has explored LLM-based agents as intermediaries in recommendation pipelines. \citet{xu2025iagent} propose \emph{iAgent}, which uses an LLM agent to extract user preference personas from interaction histories, rerank recommendations, and generate explanations, with self-reflection for iterative refinement. Their Extractor component performs a task similar to our interest prediction pipeline: converting behavioral data into structured preference representations. Our IG and IS metrics could quantify whether such agent-extracted profiles are grounded in behavioral evidence and specific enough to be actionable. Separately, \citet{ge2024scaling} demonstrate that large-scale synthetic personas can be created from diverse text sources, scaling to one billion personas via text-to-persona and persona-to-persona generation pipelines. Assigning such personas to user cohorts at the construction stage (Section~\ref{sec:dataset}) would enable evaluation of whether models can reconcile stated persona attributes with observed behavioral signals; we elaborate on this direction in Section~\ref{sec:conclusion}.

\section{Benchmark Dataset}
\label{sec:dataset}

Existing recommendation benchmarks vary significantly in their suitability for evaluating LLM-based user understanding \cite{huang2025towards, liu2025benchmarking}. To address these gaps, we construct a benchmark dataset that emulates real-world user behavior while preserving privacy through synthetic user construction. Table~\ref{tab:dataset_comparison} compares all datasets used in our benchmark across key dimensions.

\begin{table}[ht]
\centering
\small
\setlength{\tabcolsep}{3pt}
\resizebox{\columnwidth}{!}{
\begin{tabular}{l|c|c|c|c|c|c|c|c}
\toprule
\textbf{Dataset} & \textbf{Domain} & \textbf{Rich Text} & \textbf{Imp+} & \textbf{Imp--} & \textbf{Exp+} & \textbf{Exp--} & \textbf{Primary Signal} & \textbf{Density} \\
\midrule
KuaiRec~\cite{gao2022kuairec} & Short Video & Tags only & \checkmark & \checkmark & \checkmark & \ding{55} & Watch ratio & 100\% $\geq$100 eng. \\
MIND~\cite{wu2020mind} & News & \checkmark & \ding{55} & \checkmark & \checkmark & \ding{55} & Click/non-click & 39\% $\geq$100 eng. \\
Amazon Music~\cite{hou2024bridging} & Music & \checkmark & \checkmark & \ding{55} & \checkmark & \checkmark & 1--5 star ratings & 0.12\% $\geq$20 eng. \\
Goodreads~\cite{wan2018item} & Books & \checkmark & \checkmark & \ding{55} & \checkmark & \checkmark & 1--5 star ratings & Moderate \\
\midrule
\textbf{Ours (Synthetic)} & Short Video & \checkmark & \checkmark & \checkmark & \checkmark & \ding{55} & Mixed behavioral & 100\% $\geq$100 eng. \\
\bottomrule
\end{tabular}
}
\caption{Comparison of datasets used in our benchmark. \emph{Rich Text}: detailed natural language descriptions. \emph{Imp+/--}: implicit positive/negative signals. \emph{Exp+/--}: explicit positive/negative signals. \emph{Density}: fraction of users with sufficient engagement history. Each dataset presents a distinct evaluation challenge based on its signal mix and sparsity.}
\label{tab:dataset_comparison}
\end{table}

Each dataset presents distinct challenges for user understanding. KuaiRec offers dense behavioral signals but limited textual richness. MIND provides detailed news articles but lacks implicit positive signals, where articles are either clicked or not. Amazon Music and Goodreads contain explicit ratings but no implicit negative signals. Our synthetic dataset addresses these gaps by providing model-generated rich descriptions alongside the full spectrum of implicit engagement signals characteristic of modern content platforms.

\subsection{Synthetic User Construction}

To enable public release while protecting user privacy, we construct synthetic users by aggregating and anonymizing real engagement data. Our methodology ensures that no individual user can be identified while preserving the statistical properties of authentic interaction patterns.

\paragraph{Step 1: Interest-Based Cohort Formation.}
We first identify groups of users who share similar interest profiles. Specifically, we select users who have co-engaged with the same 10 interest clusters, as determined by a pre-trained interest embedding model. This ensures that users within each cohort exhibit coherent, overlapping preferences. We construct 1,000 such cohorts, requiring each cohort to contain at least 10 distinct users to provide sufficient anonymization through aggregation.

\paragraph{Step 2: Interaction History Aggregation.}
For each cohort, we aggregate the User Interaction Histories (UIHs) of all constituent users. The UIH data spans a 30-day observation window and captures the full spectrum of user engagements with video content. After aggregation, the interaction sequences are shuffled to break any temporal or user-specific patterns that could enable re-identification.

\paragraph{Step 3: Engagement Sampling.}
From the aggregated and shuffled interaction pool, we randomly sample engagements to construct each synthetic user's profile. This sampling process further obscures the provenance of individual interactions while maintaining a representative distribution of engagement types and content preferences.

\paragraph{Step 4: Content Description and Signal Categorization.}
Each sampled engagement is joined with a natural language description of the corresponding video content, generated by a proprietary Vision Language Model (VLM) that analyzes video frames and optional author metadata to produce rich textual summaries. The VLM-generated descriptions are included verbatim in the released dataset; the specific model used for generation does not affect the evaluation pipeline, as all evaluated LLMs receive identical input descriptions. These descriptions are model-generated rather than user-generated, and no direct identifiers or links to the original video content are retained. Last, we filter any model-generated text descriptions for sensitive topics and redact or synthetically replace any P.I.I.\ from the synthetic data, including names of creators.

\subsection{Interaction Signal Taxonomy}

We categorize user engagements along two orthogonal dimensions: \emph{valence} (positive vs.\ negative) and \emph{explicitness} (explicit vs.\ implicit). This taxonomy enables fine-grained analysis of how well models can distinguish between different types of user intent.

\begin{itemize}
    \item \textbf{Explicit Positive:} Deliberate actions indicating clear interest, including \emph{like}, \emph{comment}, \emph{share}, \emph{save}, \emph{follow}, and \emph{click}.
    \item \textbf{Implicit Positive:} Passive engagement signals that suggest interest without explicit action, primarily \emph{extended watch time}.
    \item \textbf{Implicit Negative:} Passive signals indicating disinterest, specifically \emph{skip} (rapid swipe to next content).
\end{itemize}

This taxonomy mirrors real-world engagement data and challenges models to appropriately weight signals of varying reliability and informativeness.

\subsection{Published Data Format}

The released dataset consists of records in the following format: each entry contains (i) a synthetic \texttt{user\_mock\_id} serving as an anonymous identifier for the synthetic user, (ii) a synthetic video summary and (iii) an interaction type label from our signal taxonomy. The dataset contains no user-generated content, no direct identifiers, and no links to original media, ensuring compliance with privacy requirements while providing a realistic testbed for User Understanding evaluation.

\subsection{Signal Characteristics and Dataset Uniqueness}

Our synthetic dataset captures the three signal types from the taxonomy: explicit positive, implicit positive, and implicit negative. A key distinguishing characteristic is the presence of implicit negative signals through skip behavior. This contrasts with explicit negative signals (e.g., low star ratings in review datasets), which represent conscious decisions to express dissatisfaction. An implicit negative signal such as a skip may instead reflect momentary disinterest, content fatigue, or simply content that does not match the user's current context~\citep{hu2008collaborative, yi2014beyond}. The presence of implicit negative signals challenges models to appropriately weight contradictory evidence, as a user who skips several cooking videos should not necessarily be classified as disinterested in cooking if they also exhibit strong positive engagement with other culinary content.

The \texttt{object\_text} field for each engagement contains a text summary in the form of hashtags of the video content, providing a natural language representation that models must parse to identify interests.

\subsection{External Datasets}

We evaluate on four established public datasets (Table~\ref{tab:dataset_comparison}), each preprocessed into a unified schema \texttt{(dataset, user\_id, object\_id, engagement\_type, object\_text, timestamp)}. KuaiRec\footnote{\url{https://kuairec.com/}} provides a nearly fully-observed interaction matrix where watch ratio is mapped to engagement types (${>}0.9$: explicit positive; $0.3$--$0.9$: implicit positive), testing inference from behavioral signals with limited textual richness. MIND\footnote{\url{https://msnews.github.io/}} offers rich article text but only binary click/non-click signals; this impoverished signal makes it ill-suited to evidence-based verification, so we exclude it from the main results and explain why in Appendix~\ref{app:mind}. Amazon Music\footnote{\url{https://cseweb.ucsd.edu/~jmcauley/datasets.html}} and Goodreads\footnote{\url{https://mengtingwan.github.io/data/goodreads}} both use 1--5 star ratings (5-star: explicit positive; 3--4: implicit positive; 1--2: explicit negative); Amazon Music exhibits extreme sparsity (87.3\% of users have only 1 review; see Appendix~\ref{app:amazon_sparsity} for implications), while Goodreads tests nuanced genre-level preference extraction.

\subsection{Dataset Validation}

We validate the fidelity of our synthetic dataset through distributional analysis comparing engagement frequencies between synthetic and real-world UIHs. Both distributions have similar right-skewed shapes with preserved relative ordering across action types. Implicit positive signals dominate, followed by implicit negative signals, and explicit positive signals are the rarest. Two-tailed t-tests show that while absolute means differ due to intentional oversampling, the relative proportions and variance structure are preserved (full distributional comparison in Figure~\ref{fig:distributions} and Table~\ref{tab:distribution_stats}, Appendix~\ref{app:distributions}). Our synthetic users differ primarily in scale, while the underlying structure of user engagement patterns is preserved. Because each synthetic user is constructed by aggregating engagements from a cohort of real users who share the same interest clusters (Section~\ref{sec:dataset}), a natural question is whether IG and IS scores computed on these aggregated profiles are comparable to scores on individual users. The survey validation in Section~\ref{sec:results} addresses this. We evaluate the same metrics on 593 individual (non-aggregated) real users and observe consistent model rankings ($\rho = 0.67$), indicating that the cohort-based construction does not distort the relative difficulty of the evaluation task. We further note that low-engagement users are filtered out during construction (Section~\ref{sec:dataset}; Table~\ref{tab:dataset_comparison}) so that every synthetic profile carries sufficient evidence for verification. This filtering accounts for the difference in scale between the aggregated-cohort and individual-user engagement distributions in Figure~\ref{fig:distributions}(a); the aggregation itself does not.

\section{Evaluation Metrics}
\label{sec:metrics}
Our evaluation framework comprises four components: (1) an \textit{Interest Prediction Pipeline} that elicits structured interest representations from LLMs, (2) a \textit{groundedness metric} that assesses whether predicted interests are supported by observable engagement evidence, decomposed into precision and recall components, (3) a \textit{specificity metric} that evaluates whether verified predicted interests are discriminative enough to identify their source content, and (4) an \textit{interest taxonomy normalization} step that enables fair cross-model comparison by mapping fine-grained interests to standardized categories before computing per-user scores.

\subsection{Interest Prediction via Structured LLM Inference}
\label{sec:prediction}

Given a user's interaction history $\mathcal{H}_u$ containing content objects with engagement signals (e.g., likes, watch time, skips), the LLM generates a User Interest Profile $\mathcal{I}_u = \{(I_j, \mathcal{O}_j)\}_{j=1}^{m}$, where each interest $I_j$ is a natural language descriptor grounded in a set of cited evidence objects $\mathcal{O}_j \subseteq \mathcal{H}_u$. Engagement signals are categorized along two dimensions: \textit{valence} (positive vs.\ negative) and \textit{explicitness} (explicit actions like likes vs.\ implicit signals like watch time). Not all datasets contain all signal types; our pipeline adapts to each dataset's available taxonomy.

\paragraph{Programmatic Instruction Construction.}
A key design principle is \textit{closed-loop verification}: the exact evidence thresholds used to verify interests (Section~\ref{sec:ig}) are programmatically encoded into the generation prompt, establishing a 1-to-1 correspondence between instruction and evaluation. For each dataset, we translate formal verification predicates (e.g., ``require $\geq$3 implicit positive signals'') into natural language constraints embedded in the prompt. This ensures that verification failures reflect genuine model limitations (insufficient evidence discovery or hallucination) rather than misalignment between generation expectations and evaluation criteria. The prompt instructs the model to: (i) produce specific interests (2--5 words) rather than generic categories, (ii) cite sufficient positive evidence while respecting negative signal constraints, and (iii) associate each object with at most two interests to encourage specificity.

\subsection{Groundedness}
\label{sec:ig}
In the absence of ground-truth, we rely on evidence-based verification: predicted interests are checked against observable engagement signals in the user's interaction history rather than subjective labels or model self-assessment.
The core insight behind this is that a credible interest prediction should not be based on a single, potentially noisy engagement signal. Rather, it should be corroborated by multiple independent pieces of evidence from the user's interaction history. The design principle is that claims require sufficient supporting evidence to be considered verified. Our verification scheme operationalizes this principle by defining configurable evidence thresholds that predicted interests must satisfy.

Formally, let $\mathcal{I}_u = \{I_1, \ldots, I_m\}$ denote the set of interests predicted for user $u$, and let $\mathcal{O}_j \subseteq \mathcal{H}_u$ denote the set of evidence objects cited for interest $I_j$. For each evidence object $o \in \mathcal{O}_j$, let $e(o) \in \mathcal{E}$ denote its engagement type. We partition the engagement space into positive signals $\mathcal{E}^{+}$ (indicating affinity) and negative signals $\mathcal{E}^{-}$ (indicating disinterest), each further subdivided into explicit and implicit categories. We define the counts:
\begin{align}
n^+_{\text{exp}}(I_j) &= |\{o \in \tilde{\mathcal{O}}_j : e(o) \in \mathcal{E}^{+}_{\text{exp}}\}| \\
n^+_{\text{imp}}(I_j) &= |\{o \in \tilde{\mathcal{O}}_j : e(o) \in \mathcal{E}^{+}_{\text{imp}}\}| \\
n^-_{\text{exp}}(I_j) &= |\{o \in \tilde{\mathcal{O}}_j : e(o) \in \mathcal{E}^{-}_{\text{exp}}\}| \\
n^-_{\text{imp}}(I_j) &= |\{o \in \tilde{\mathcal{O}}_j : e(o) \in \mathcal{E}^{-}_{\text{imp}}\}|
\end{align}
where $\tilde{\mathcal{O}}_j$ is the judge-filtered evidence set (defined below) and $\mathcal{E}^{+}_{\text{exp}}$, $\mathcal{E}^{+}_{\text{imp}}$, $\mathcal{E}^{-}_{\text{exp}}$, $\mathcal{E}^{-}_{\text{imp}}$ are the sets of explicit positive, implicit positive, explicit negative, and implicit negative engagement types, respectively.

\paragraph{Evidence Filtering via LLM Judge.}
Before applying verification thresholds, we filter the cited evidence set to ensure semantic relevance. For each interest $I_j$, an LLM judge (Llama-3.3-70B-Instruct~\cite{llama3}, the same model used for IS evaluation in Section~\ref{sec:is}) independently evaluates every cited object $o \in \mathcal{O}_j$ and determines whether it is genuinely relevant to the predicted interest. Let $\tilde{\mathcal{O}}_j \subseteq \mathcal{O}_j$ denote the filtered subset of objects deemed relevant by the judge. The signal counts above are then computed over $\tilde{\mathcal{O}}_j$ rather than $\mathcal{O}_j$. Without this step, models could inflate their signal counts by citing objects that share superficial lexical similarity with an interest but lack genuine topical relevance. The judge decouples the generation model's citation behavior from the verification outcome, so that only semantically grounded evidence contributes to the signal counts.

An interest $I_j$ is considered \textit{verified}, denoted $\textsc{Verified}(I_j) = 1$, if it satisfies a dataset-specific verification predicate $\phi_{\mathcal{D}}$ that encodes two complementary constraints:
\begin{equation}
\textsc{Verified}(I_j) = \phi_{\mathcal{D}}\left(n^+_{\text{exp}}, n^+_{\text{imp}}, n^-_{\text{exp}}, n^-_{\text{imp}}\right)
\end{equation}

The asymmetric thresholds reflect the differential reliability of signal types. \textit{Explicit signals} (likes, saves, shares) represent deliberate user actions with high signal-to-noise ratio~\citep{hu2008collaborative}, which justifies a lower threshold ($\geq$2). \textit{Implicit signals} (watch time) are inherently noisier, as extended viewing may reflect distraction or auto-play rather than genuine interest~\citep{yi2014beyond}, which requires stronger corroboration ($\geq$3). The hybrid threshold (1 explicit + 2 implicit) captures cases where deliberate action is supported by sustained passive engagement.

Negative constraints are set conservatively: up to 3 implicit negatives (skips) or 2 explicit negatives (dislikes) are tolerated before invalidating an interest. This acknowledges that negative signals are often context-dependent (e.g. users may skip content due to timing or mood rather than fundamental disinterest). Only interests with \emph{consistent} contradictory evidence are filtered. These thresholds reflect domain-informed design choices: explicit signals require only $\geq$2 because they are high-precision, while implicit signals require $\geq$3 to compensate for noise. We validate this configuration against user self-reported interests (Table~\ref{tab:survey_validation}), achieving $\rho = 0.67$ correlation between our metrics and survey-based ground truth, and conduct a formal threshold sensitivity analysis (Appendix~\ref{app:threshold}) confirming that model rankings remain stable under $\pm1$ perturbations of every threshold. In a production recommendation system, these thresholds encode a deployment requirement: the minimum evidence standard for an interest to be considered actionable. The benchmark evaluates whether models can satisfy these requirements. The thresholds themselves are a domain-specific design choice, not a methodological assumption, and can be recalibrated for different deployment contexts.

The verification predicate $\phi_{\mathcal{D}}$ is composed of two components: (1) a \textit{positive evidence requirement} that ensures the interest is supported by sufficient affirmative engagement signals, and (2) a \textit{negative evidence constraint} that ensures the interest is not contradicted by excessive disengagement signals. This dual-criterion structure allows us to verify not only that an interest has adequate supporting evidence, but also that this evidence is not undermined by contradictory signals from the same user.

The specific instantiation of $\phi_{\mathcal{D}}$ is configurable based on the engagement types available in each dataset, ensuring generalization across heterogeneous recommendation domains. As established in Section~\ref{sec:prediction}, these predicates are communicated to the model during generation via closed-loop instruction construction.

Each interest thus yields a binary verification outcome $\textsc{Verified}(I_j) \in \{0, 1\}$. The aggregation of these per-interest outcomes into a final score, including normalization across interest categories, is described in Section~\ref{sec:aggregation}.

 This metric is conceptually aligned with the notion of \textit{faithfulness} in explainable AI~\citep{jacovi2020towards}: a prediction is faithful if it accurately reflects the underlying evidence. By interposing an LLM judge to filter citations for relevance, then requiring that the filtered evidence satisfies configurable signal thresholds, we penalize both hallucinated interests and superficially grounded predictions that lack genuine behavioral support.

\subsection{Specificity}
\label{sec:is}

While the groundedness metric (Section~\ref{sec:ig}) assesses whether predicted interests are supported by sufficient evidence, it does not evaluate whether those interests are \textit{specific} enough to meaningfully characterize the user. A model could achieve high verification scores by predicting overly generic interests (e.g., ``Entertainment'') that trivially apply to many engagement signals. To address this, we introduce a complementary \textit{specificity metric} that tests whether predicted interests are sufficiently granular to identify their source content from a pool of distractors. The intuition is that a well-specified interest should exhibit strong mutual information with the content that generated it. We operationalize this via a retrieval task: given an interest and a set of candidate objects (most of which are unrelated), can an evaluator correctly identify which objects support the interest?

Formally, for each user $u$ and predicted interest $I_j \in \mathcal{I}_u$, let $\mathcal{O}_j = \{o_1, \ldots, o_c\}$ denote the set of $c$ evidence objects cited by the model for this interest. The specificity metric measures the ability of an LLM judge to recover these evidence objects from a mixed test set containing both correct (cited) and incorrect (distractor) objects.

\paragraph{Test Set Construction.}
For each interest $I_j$, we construct a test set $\mathcal{T}_j$ of size $|\mathcal{T}_j| = N$ (we use $N=50$) as follows:

\begin{enumerate}
    \item \textbf{Evidence sampling}: Select up to $n = \min(|\mathcal{O}_j|, 5)$ evidence objects from $\mathcal{O}_j$. Capping at 5 prevents interests with many citations from dominating.
    \item \textbf{Global pool construction}: Construct a global pool $\mathcal{P}$ of objects sampled uniformly across all users in the dataset, with $|\mathcal{P}| = 1000$. For each object, we record all interests it has been cited for across users.
    \item \textbf{Overlap filtering}: To avoid false negatives, we query an LLM to identify and remove objects from $\mathcal{P}$ whose associated interests semantically overlap with those of user $u$ (e.g., ``football'' overlaps with ``soccer''). Let $\mathcal{P}_u \subseteq \mathcal{P}$ denote the filtered pool.
    \item \textbf{Distractor sampling}: Sample $N - n$ distractor objects uniformly from $\mathcal{P}_u$.
    \item \textbf{Shuffling}: Combine the $n$ evidence objects with the $N - n$ distractors and randomly shuffle to form $\mathcal{T}_j$.
\end{enumerate}

\paragraph{Evaluation Protocol.}
For each interest $I_j$, we present an LLM judge with: (i) the interest text, (ii) the number of backing videos $n$, and (iii) textual descriptions of all $N$ objects in $\mathcal{T}_j$ with anonymized labels. The judge is instructed to identify exactly $n$ objects that most likely led to the inference of this interest.

\paragraph{LLM Judge Configuration.}
For this specificity evaluation, we employ Llama-3.3-70B-Instruct as the judge model. The judge operates with temperature 0 to ensure deterministic evaluations across runs. Each prompt presents the judge with the interest text, the exact number of backing videos $n$ to identify, and anonymized video descriptions (labeled \texttt{video\_0}, \texttt{video\_1}, etc.) shuffled to prevent position bias. The system prompt instructs the judge to act as ``an expert in content analysis and user interest understanding'' and to output only comma-separated video identifiers. To ensure independence between the interest generation and evaluation phases, we enforce strict separation: the judge receives only the predicted interest text and candidate object descriptions, with no access to the original user interaction history or the generation model's internal reasoning. The judge model (Llama-3.3-70B-Instruct) is distinct from all evaluated models, differing in architecture version or model family.

Let $\hat{\mathcal{O}}_j$ denote the set of objects selected by the judge. For each interest $I_j$, the evaluation produces two counts: $\textsc{correct}(I_j) = |\hat{\mathcal{O}}_j \cap \mathcal{O}_j|$, the number of correctly identified evidence objects, and $\textsc{backing}(I_j) = n$, the total number of true evidence objects in the test set. These per-interest counts are aggregated into a final score via interest taxonomy normalization, as described in Section~\ref{sec:aggregation}.

This metric is conceptually aligned with \textit{plausibility} in the faithfulness-plausibility framework~\citep{jacovi2020towards}: an interest is plausible if it is specific enough to be meaningfully associated with particular content. Overly generic interests may appear superficially reasonable but fail the specificity test because they cannot discriminate between their true evidence and unrelated distractors. Together, the groundedness and specificity metrics reward profiles that are both faithful (supported by evidence) and plausible (sufficiently distinctive).

\subsection{Interest Category Normalization}
\label{sec:normalization}

The per-interest metrics defined in Sections~\ref{sec:ig} and~\ref{sec:is} produce raw verification flags and specificity counts at the level of individual predicted interests. However, directly aggregating across interests conflates \textit{depth} (many predictions within a topic) with \textit{breadth} (coverage across diverse topics). For example, a model that predicts ``LeBron James'', ``NBA Draft Picks'', and ``Lakers Trade Rumors'' (all mapping to Basketball) would be rewarded three times for a single topical area. By first grouping fine-grained interests into standardized interest categories and computing per-category ratios, all three basketball predictions contribute a single ratio (e.g., $3/3 = 1.0$), while breadth across distinct categories is appropriately rewarded.

We map each predicted fine-grained interest to a broader \textit{interest category} drawn from a hierarchical taxonomy comprising 35 domains and 325 categories (Table~\ref{tab:taxonomy-examples}). Each category is defined by explicit labeling guidelines specifying its scope and boundaries. Categories span both topical interests (e.g., Basketball, Artificial Intelligence) and content formats (e.g., Vlogs, Street Interviews). Any comparable taxonomy with sufficient granularity could serve the same purpose.

  \begin{table}[t]
  \centering
  \small
  \caption{Examples of top-level interest domains and their
  corresponding interest categories. Each domain contains several fine-grained interest categories; only a
  representative subset is shown. The full list of
  categories is provided in Appendix~\ref{app:taxonomy}.}
  \label{tab:taxonomy-examples}
  \begin{tabular}{@{}p{0.22\columnwidth}p{0.72\columnwidth}@{}}
  \toprule
  \textbf{Category Description} & \textbf{Example Interest Categories} \\
  \midrule
  Sports & Basketball, American Football, Soccer / Football,
  Combat Sports \& Martial Arts, Golf, Winter Sports,
  \ldots \\[4pt]
  Food \& Cooking & Baked Goods \& Dessert, Food Recipes \&
  Cooking Tips, Local / Regional Cuisine, Mukbang,
  Restaurants \& Eateries, \ldots \\[4pt]
  Electronics \& Technology & Artificial Intelligence,
  Software \& Apps, Computers \& Electronic Devices,
  Tech Builds \& Repairs, \ldots \\[4pt]
  Comedy \& Humor & Stand Up \& Roasting, Sketches \& Skits,
  Pranks, Memes, Reaction Comedy, \ldots \\[4pt]
  Visual Arts \& Design & Drawing \& Painting, Photography,
  Graphic Design, Sculpture, Street \& Public Art,
  \ldots \\
  \bottomrule
  \end{tabular}
  \end{table}

The taxonomy was originally developed for video content classification. Its video-centric origin means some categories (e.g., Mukbang, Street Interviews) are less relevant for non-video datasets; however, its broad topical coverage and explicit category definitions make it well-suited as a normalization layer. As noted above, any comparable taxonomy with sufficient granularity could serve the same purpose; the full category list is provided in Appendix~\ref{app:taxonomy} to enable replication with alternative taxonomies.

\paragraph{Free-Form Interest to Taxonomy Mapping.}
We extract globally unique fine-grained interests across all models and users, ensuring identical interests are mapped consistently regardless of which model predicted them. An LLM classifies each interest into exactly one category, processing interests in batches with the full category list as context. Mappings are manually audited by the authors for accuracy.

\paragraph{Why Taxonomy Normalization Is Necessary.}
Beyond preventing depth-breadth conflation, the taxonomy serves a more fundamental role: it enables verification in the absence of ground truth. Without a shared categorization, semantically near-identical predictions (e.g., ``NBA Trade Rumors'' and ``NBA Free Agency News'') would be counted as independent successes, inflating scores for models that produce redundant paraphrases of the same underlying interest. By collapsing such predictions into a single category, the taxonomy ensures that verification reflects genuine breadth of understanding rather than lexical diversity. This normalization also enables the construction of an \textit{ensemble oracle}: aggregating verified interest categories across multiple models to identify a consensus set of ground-truth interests per user, without requiring human annotation. When diverse model architectures independently converge on the same interest category, confidence in that interest's validity increases. We adopt this approach for computing IG Recall (Section~\ref{sec:aggregation}).

\subsection{Score Aggregation}
\label{sec:aggregation}

We compute per-user IG and IS scores via a two-step process: (1) computing per-category ratios using the taxonomy mapping from Section~\ref{sec:normalization}, and (2) aggregating these ratios through a precision-recall framework for groundedness and verified-only averaging for specificity.

\paragraph{Per-Category Ratios.}
Let $\mathcal{C}_u$ denote the set of interest categories to which user $u$'s predicted interests are mapped. For each category $c \in \mathcal{C}_u$, let $\mathcal{I}_c = \{I_j : c(I_j) = c\}$ denote the subset of interests assigned to $c$. We define the per-category groundedness ratio as the fraction of interests in $c$ that are verified:
\begin{equation}
G_c = \frac{\sum_{I_j \in \mathcal{I}_c} \textsc{Verified}(I_j)}{|\mathcal{I}_c|}
\end{equation}
and the per-category specificity ratio as the fraction of correctly identified evidence objects across all interests in $c$:
\begin{equation}
S_c = \frac{\sum_{I_j \in \mathcal{I}_c} \text{correct}(I_j)}{\sum_{I_j \in \mathcal{I}_c} \text{backing}(I_j)}
\end{equation}
Computing ratios within each interest category before aggregating across categories ensures that a cluster of related interests (e.g., multiple sports-related predictions mapping to the same category) contributes a single ratio rather than inflating the score through redundancy.

\paragraph{Groundedness: Precision-Recall Decomposition.}
Aggregating per-category groundedness ratios requires a normalization denominator. Using a single denominator conflates two distinct dimensions: \textit{precision} (are predicted categories genuine?) and \textit{recall} (are all discoverable categories found?). We therefore decompose IG into precision and recall components.

We define the \textit{oracle count} as the number of interest categories containing at least one verified interest across \emph{any} evaluated model for user $u$:
\begin{equation}
\text{Oracle}_u = \bigl|\{c \in \textstyle\bigcup_M \mathcal{C}_{u}^{(M)} : \exists\, I_j \in \mathcal{I}_c \text{ with } \textsc{Verified}(I_j) = 1\}\bigr|
\end{equation}
The oracle represents the set of discoverable interest categories for user $u$ according to the collective evidence of all models, providing a model-independent upper bound on verifiable breadth.

\textit{IG Precision} measures the fraction of predicted categories that are verified, penalizing hallucinated categories that contribute zero verified interests:
\begin{equation}
\text{IG}_P = \frac{\sum_{c \in \mathcal{C}_u} G_c}{|\mathcal{C}_u|}
\end{equation}
\textit{IG Recall} measures the fraction of oracle categories that the model successfully covers:
\begin{equation}
\text{IG}_R = \frac{\sum_{c \in \mathcal{C}_u} G_c}{\text{Oracle}_u}
\end{equation}

The primary groundedness score is their harmonic mean:
\begin{equation}
\text{IG}_{F1} = \frac{2 \cdot \text{IG}_P \cdot \text{IG}_R}{\text{IG}_P + \text{IG}_R}
\end{equation}

This decomposition reveals complementary failure modes: a model predicting many unverified categories achieves low IG$_P$ despite potentially high IG$_R$ (it finds real interests but also hallucinates), while a conservative model predicting few but verified categories achieves high IG$_P$ but low IG$_R$ (precise but incomplete).

\paragraph{Specificity: Verified-Only Averaging.}
The specificity metric should evaluate description quality only for categories where the model has demonstrated genuine understanding. Averaging $S_c$ over unverified categories, where the interest itself may be hallucinated, would conflate description quality with hallucination, and can cause IS to exceed 1.0. We therefore restrict IS to verified categories:
\begin{equation}
\text{IS}_u = \frac{\sum_{c \in \mathcal{V}_u} S_c}{|\mathcal{V}_u|}, \quad \mathcal{V}_u = \{c \in \mathcal{C}_u : G_c > 0\}
\end{equation}
where $\mathcal{V}_u$ is the set of categories containing at least one verified interest for the current model. This ensures IS is bounded in $[0, 1]$ and measures the discriminative specificity of genuine interests only. When $\mathcal{V}_u = \emptyset$ (no verified interests), IS$_u$ is defined as 0.

We report IG$_{F1}$ and IS as the primary benchmark scores, with IG$_P$ and IG$_R$ providing diagnostic insight into model behavior. Scores are reported as the median across all evaluated users.

\subsection{Metric Computation Pipeline}
\label{sec:pipeline}

The end-to-end evaluation pipeline transforms raw LLM generations into final per-user scores through three stages, summarized below.

\paragraph{Stage 1: Per-Interest Evaluation.}
Raw LLM generations are parsed to extract interest-object mappings. For each interest, the groundedness evaluation (Section~\ref{sec:ig}) applies an LLM judge to filter cited objects for semantic relevance, then applies dataset-specific verification predicates to the filtered evidence, producing a binary verification flag per interest. In parallel, the specificity evaluation (Section~\ref{sec:is}) constructs a test set of evidence and distractor objects, presents it to an LLM judge, and records the number of correctly identified evidence objects per interest.

\paragraph{Stage 2: Interest Category Mapping.}
Globally unique interests across all models are extracted and mapped to interest categories from the standardized taxonomy (Section~\ref{sec:normalization}). This mapping is computed once and shared across all models and users to ensure consistency.

\paragraph{Stage 3: Score Aggregation.}
Per-interest verification flags and specificity counts from Stage~1 are grouped by their category assignments from Stage~2. Per-category groundedness ratios are aggregated via the precision-recall decomposition (IG$_P$, IG$_R$, IG$_{F1}$), and per-category specificity ratios are averaged over verified categories only to produce IS (Section~\ref{sec:aggregation}). The overall benchmark scores are reported as the median across users. We quantify the token and GPU cost of running this pipeline in Appendix~\ref{app:compute_cost}.

\section{Experimental Setup}
\label{sec:setup}

\paragraph{UIH Segmentation.}
For users with large interaction histories, we segment the UIH into chunks of at most 100 engagements per prompt. Engagements are first sorted by recency (most recent first), then partitioned into non-overlapping windows. Each window is processed as an independent prompt, and the resulting interest predictions are aggregated at the user level. This chunking strategy ensures that prompts remain within context limits while preserving temporal structure.

\paragraph{Prompt Construction.}
Each engagement record is formatted as a numbered entry containing the interaction type and object description. The prompt instructs the model to identify interests grounded in sufficient evidence, with dataset-specific thresholds encoded directly in the instruction (see Section~\ref{sec:ig}). Each object is assigned a sequential identifier that the model references when citing evidence, enabling reliable parsing and mapping to source objects.

\paragraph{Models and Inference.}
We evaluate eight open-weight LLMs spanning different model families, scales, and training paradigms (Table~\ref{tab:model_ids}, Appendix~\ref{app:model_ids}): Llama-3.1-8B-Instruct~\cite{llama3}, Qwen2-7B-Instruct~\cite{qwen2}, Qwen2.5-72B-Instruct~\cite{qwen2.5}, Qwen3-32B and Qwen3-VL-235B-A22B-Instruct~\cite{qwen3}, DeepSeek-R1-Distill-Llama-70B and DeepSeek-R1~\cite{deepseekr1}, and GPT-OSS-120B~\cite{gptoss}. We also evaluate three proprietary frontier models (GPT-5, Claude 4.6, and Gemini 3.5 Flash) through the same generation and metric harness, accessing them through a hosted API rather than local serving. The open-weight models were served on NVIDIA H100 80GB GPUs using 8-way tensor parallelism via a proprietary inference framework. No weight quantization was applied (full bf16 precision). For reproducibility, all inference uses temperature 0 with top-p 0. The maximum context length is set to approximately 100K tokens, with generation limited to 6,048 tokens per call. Batch size was 256. The IG evidence-filtering judge and IS retrieval judge both used Llama-3.3-70B-Instruct (Section~\ref{sec:is}). The eight primary models are open-weight; results are reproducible with any compatible serving framework given the same generation parameters. We sample up to 1,000 users per dataset using deterministic hashing on user identifiers to ensure reproducibility across runs; the same hashing scheme is applied to object identifiers for deterministic distractor pool construction. All intermediate outputs undergo best-effort JSON extraction: we first attempt standard parsing, then apply regex-based extraction to recover JSON objects embedded in free-text responses, and finally use automated JSON repair to fix common structural errors (unclosed brackets, trailing commas). Generations that remain unparsable after these recovery steps are filtered before metric computation.

\section{Results}
\label{sec:results}
Table~\ref{tab:merged_results} reports median IG$_{F1}$ and median IS scores across users for each model and dataset. Best results per dataset are \textbf{bolded}.
\begin{table}[ht]
\centering
\small
\setlength{\tabcolsep}{6pt}
\resizebox{\columnwidth}{!}{
\begin{tabular}{l|cc|cc|cc|cc}
\toprule
& \multicolumn{2}{c|}{\textbf{Synthetic}} & \multicolumn{2}{c|}{\textbf{KuaiRec}} & \multicolumn{2}{c|}{\textbf{Amazon Music}} & \multicolumn{2}{c}{\textbf{Goodreads}} \\
\textbf{Model} & IG$_{F1}$ & IS & IG$_{F1}$ & IS & IG$_{F1}$ & IS & IG$_{F1}$ & IS \\
\midrule
GPT-OSS-120B          & \textbf{67.8} & 74.0 & 56.3 & 68.4 & 66.7 & 70.6 & 58.0 & 67.4 \\
GPT-5                 & 59.7 & 81.1 & 62.3 & 74.4 & 66.7 & \textbf{90.9} & \textbf{70.0} & 83.7 \\
DeepSeek-R1           & 56.5 & 79.0 & 44.1 & 71.0 & 57.1 & 74.1 & 43.5 & 68.9 \\
Qwen3-VL-235B         & 50.8 & 73.4 & 55.2 & 65.9 & 50.0 & 70.8 & 48.5 & 65.5 \\
Claude 4.6            & 50.1 & 78.0 & \textbf{66.9} & 69.7 & \textbf{75.0} & 86.8 & 61.3 & 83.4 \\
Gemini 3.5 Flash      & 47.8 & \textbf{83.6} & 58.4 & \textbf{78.3} & 66.7 & 90.0 & 45.3 & \textbf{87.9} \\
Qwen3-32B             & 46.1 & 74.5 & 45.8 & 66.0 & 50.0 & 73.3 & 40.1 & 62.1 \\
Qwen2.5-72B           & 45.5 & 65.1 & 43.3 & 65.2 & 43.3 & 60.0 & 42.1 & 57.0 \\
\makecell[l]{DeepSeek-R1-\\Distill-70B} & 37.3 & 78.0 & 42.9 & 69.2 & 33.3 & 75.0 & 33.3 & 66.5 \\
Llama-3.1-8B-Instruct & 15.9 & 33.0 & 30.6 & 56.0 & 37.3 & 25.3 & 19.1 & 19.9 \\
Qwen2-7B              & 14.2 & 31.8 & 26.5 & 22.0 & 28.6 & 40.0 & 3.2 & 0.0 \\
\bottomrule
\end{tabular}
}
\caption{Median IG$_{F1}$ (\%) and IS (\%) across the four primary benchmark datasets. IG$_{F1}$ is the harmonic mean of IG Precision and IG Recall; IS is computed over verified interest categories only (bounded $[0,1]$). Best per-dataset results in \textbf{bold}.}
\label{tab:merged_results}
\end{table}

No single model dominates: GPT-OSS-120B achieves the best IG$_{F1}$ on Synthetic, Claude 4.6 on KuaiRec and Amazon Music, and GPT-5 on Goodreads, while Gemini 3.5 Flash leads IS on three of four datasets, reflecting a precision-recall trade-off. DeepSeek-R1 predicts fewer but more precise categories (high IG$_P$) and produces highly specific descriptions for its verified interests. Dataset difficulty is a major source of variation; see Section~\ref{sec:dataset} for a discussion of Amazon Music sparsity. IS is bounded $[0, 1]$ by construction. Models with zero verified categories (IG$_{F1} = 0$) correctly report IS $= 0$, so unverified categories cannot inflate specificity scores. Figure~\ref{fig:ig_pr_benchmark} (Appendix~\ref{app:per_user}) shows the per-user score distributions underlying these medians.

Table~\ref{tab:ig_pr_breakdown} decomposes IG$_{F1}$ into its precision and recall components on the Synthetic and KuaiRec datasets. Across all models, IG$_P >$ IG$_R$: models are generally more precise than they are complete, and the bottleneck is \emph{coverage} rather than hallucination. DeepSeek-R1 shows this pattern most strongly. It achieves 82.4\% IG$_P$ on Synthetic (the highest) but only 43.9\% IG$_R$, indicating that nearly all of its predicted categories are genuine, yet it discovers less than half of all discoverable interests.

\begin{table}[t]
\centering
\small
\setlength{\tabcolsep}{3pt}
\begin{tabular}{l|cc|cc}
\toprule
& \multicolumn{2}{c|}{\textbf{Synthetic}} & \multicolumn{2}{c}{\textbf{KuaiRec}} \\
\textbf{Model} & IG$_P$ & IG$_R$ & IG$_P$ & IG$_R$ \\
\midrule
DeepSeek-R1           & \textbf{82.4} & 43.9 & \textbf{89.8} & 29.5 \\
GPT-OSS-120B          & 78.1 & \textbf{60.5} & 79.7 & 44.0 \\
Qwen3-32B             & 73.6 & 34.3 & 82.4 & 31.8 \\
GPT-5                 & 67.8 & 53.5 & 73.5 & 54.2 \\
Qwen2.5-72B           & 65.7 & 34.8 & 79.2 & 29.7 \\
Gemini 3.5 Flash      & 63.6 & 38.4 & 84.2 & 44.9 \\
Qwen3-VL-235B         & 57.5 & 45.3 & 74.0 & 44.5 \\
\makecell[l]{DeepSeek-R1-\\Distill-70B} & 54.9 & 28.8 & 71.8 & 31.1 \\
Claude 4.6            & 50.0 & 49.1 & 79.2 & \textbf{57.5} \\
Qwen2-7B              & 28.9 & 9.4 & 53.5 & 17.9 \\
Llama-3.1-8B-Instruct          & 20.8 & 12.9 & 50.0 & 21.8 \\
\bottomrule
\end{tabular}
\caption{IG Precision and IG Recall (\%) on Synthetic and KuaiRec. IG$_P >$ IG$_R$ for every model. The primary bottleneck is coverage; hallucination is not. Best IG$_P$ and IG$_R$ per dataset in \textbf{bold}.}
\label{tab:ig_pr_breakdown}
\end{table}

\subsection{Validation Against User Surveys}

To validate our metrics against user-reported ground truth, we compare IG and IS with Survey Precision (S-Prec) and Survey Recall (S-Rec) computed by matching model-predicted interests against self-reported interest labels (Table~\ref{tab:survey_validation}). This validation uses 593 users who self-reported their interests via an in-product survey. The survey instrument and recruitment details are proprietary; however, the validation methodology itself (matching model-predicted interests against user-stated labels) is dataset-agnostic and can be replicated with any population that provides self-reported interest labels. S-Prec/S-Rec measure alignment with \emph{user-stated identity}, while IG/IS measure \emph{behavioral grounding} and \emph{discriminative specificity} from engagement signals. The Spearman correlation between survey F1 and the geometric mean of IG$_{F1}$ and IS is $\rho = 0.67$: the two metric families broadly agree on model ranking but diverge on specific models. The precision-recall decomposition reveals that DeepSeek-R1 achieves the highest IG$_P$ (57.3\%) and IG$_{F1}$ (42.9\%), confirming its predicted categories are both genuine and well-covered. In contrast, Qwen3-VL-235B achieves the highest S-Rec but the lowest IG$_P$ (25.0\%) among capable models: it discovers many interest categories but over half are unverified, explaining its moderate IG$_{F1}$ despite high coverage. This confirms that IG/IS capture complementary facets of user understanding that are not redundant with alignment to conscious user identity. Figure~\ref{fig:ig_pr_survey} (Appendix~\ref{app:ig_pr_survey}) visualizes this precision-recall tradeoff with iso-F1 contours.

\begin{table}[t]
\centering
\small
\setlength{\tabcolsep}{2pt}
\begin{tabular}{l|cc|cccc}
\toprule
& \multicolumn{2}{c|}{\textbf{Survey-Based}} & \multicolumn{4}{c}{\textbf{Proposed Metrics}} \\[0.3em]
\textbf{Model} & S-Prec & S-Rec & IG$_P$ & IG$_R$ & IG$_{F1}$ & IS \\
\midrule
DeepSeek-R1           & 62.3 & 60.0 & 57.3 & 33.7 & 42.9 & 83.3 \\
GPT-OSS-120B          & 59.0 & 60.2 & 31.6 & 26.5 & 27.7 & 87.8 \\
Qwen3-32B             & 59.5 & 52.8 & 50.0 & 26.1 & 33.9 & 80.6 \\
Qwen3-VL-235B         & 57.1 & 66.1 & 25.0 & 26.9 & 25.7 & 84.8 \\
\makecell[l]{DeepSeek-R1-\\Distill-70B} & 51.4 & 56.1 & 25.2 & 21.4 & 24.3 & 83.3 \\
Qwen2.5-72B           & 49.2 & 56.6 & 46.4 & 32.7 & 37.7 & 76.7 \\
Llama-3.1-8B-Instruct & 46.0 & 43.0 & 16.0 & 10.0 & 12.3 & 40.0 \\
Qwen2-7B              & 43.1 & 49.5 & 19.3 & 12.5 & 14.9 & 46.7 \\
\midrule
\multicolumn{7}{l}{Spearman $\rho$(Survey F1, $\sqrt{\text{IG}_{F1} \cdot \text{IS}}$) $= 0.67$} \\
\bottomrule
\end{tabular}
\caption{Survey-Based metrics (S-Prec, S-Rec) versus proposed metrics (IG$_P$, IG$_R$, IG$_{F1}$, IS), all in \%. S-Prec and S-Rec are computed against user self-reported interests; IG and IS are computed from engagement signals (Section~\ref{sec:metrics}). The full precision-recall decomposition reveals that IG$_P >$ IG$_R$ for seven of the eight models on the survey dataset (Qwen3-VL-235B is the recall-leaning exception), consistent with the benchmark results (Table~\ref{tab:ig_pr_breakdown}): coverage, not hallucination, is the bottleneck. $\rho$: Spearman correlation between survey F1 and geometric mean of IG$_{F1}$ and IS.}
\label{tab:survey_validation}
\end{table}

\subsection{Key Findings}

\paragraph{Finding 1: Instruction Following Is Necessary but Does Not Predict Rank.}
Table~\ref{tab:instruction_following} shows the relationship between instruction-following capability, measured by the Chatbot Arena Instruction Following Elo rating~\cite{chiang2024chatbot}, and median IG$_{F1}$ on the survey dataset.

\begin{table}[t]
\centering
\small
\setlength{\tabcolsep}{4pt}
\begin{tabular}{l|c|c}
\toprule
\textbf{Model} & \textbf{Arena IF Elo} & \textbf{Median IG$_{F1}$ (\%)} \\
\midrule
DeepSeek-R1        & 1395 & 42.9 \\
Qwen2.5-72B        & 1238 & 37.7 \\
Qwen3-32B          & 1330 & 33.9 \\
GPT-OSS-120B       & 1325 & 27.7 \\
Qwen3-VL-235B      & 1325 & 25.7 \\
\makecell[l]{DeepSeek-R1-\\Distill-70B} & 1395 & 24.3 \\
Qwen2-7B           & 1200 & 14.9 \\
Llama-3.1-8B-Instruct & 1189 & 12.3 \\
\bottomrule
\end{tabular}
\caption{Instruction-following capability (Chatbot Arena IF Elo~\cite{chiang2024chatbot}) and median IG$_{F1}$ (\%) on the survey dataset (real users). Among these models, Elo does not predict IG$_{F1}$ rank.}
\label{tab:instruction_following}
\end{table}

Arena IF Elo correlates moderately with IG$_{F1}$ across the full model set (Spearman $\rho = 0.55$, $n = 8$, computed from Table~\ref{tab:instruction_following}), but the correlation is driven by a floor effect. Models with Elo at or below $\sim$1200 score lowest: Llama-3.1-8B-Instruct (Elo 1189) reaches only 12.3\% median IG$_{F1}$, and Qwen2-7B (Elo 1200) only 14.9\%. Above this threshold, Elo loses predictive power. Among the six models with Elo $> 1200$, Spearman $\rho = -0.1$ ($n = 6$). DeepSeek-R1-Distill-70B shares R1's Elo rating (1395) but achieves only 24.3\% IG$_{F1}$ versus 42.9\%. Qwen2.5-72B has the \emph{lowest} Elo among capable models (1238) but the second-highest IG$_{F1}$ (37.7\%). Once format compliance is met, domain-specific reasoning about engagement signals, not general instruction compliance, determines model ranking.

\paragraph{Finding 2: Evidence Counting Is the Primary Bottleneck.}
Table~\ref{tab:verification_failures} breaks down why interests fail the IG criteria.

\begin{table}[t]
\centering
\small
\setlength{\tabcolsep}{4pt}
\begin{tabular}{l|c|c|c}
\toprule
\textbf{Model} & \textbf{Ins. Imp+} & \textbf{Ins. Exp+} & \textbf{Exc. Neg} \\
\midrule
DeepSeek-R1           & 86.7 & 99.4 & 1.8 \\
GPT-OSS-120B          & 97.3 & 99.2 & 1.3 \\
Qwen3-32B             & 95.7 & 98.6 & 3.1 \\
Qwen3-VL-235B         & 93.1 & 98.3 & 10.8 \\
\makecell[l]{DeepSeek-R1-\\Distill-70B} & 90.5 & 94.8 & 9.6 \\
Qwen2.5-72B           & 76.3 & 92.0 & 22.9 \\
Llama-3.1-8B-Instruct & 95.8 & 97.8 & 10.4 \\
Qwen2-7B              & 73.6 & 91.8 & 22.0 \\
\bottomrule
\end{tabular}
\caption{Distribution of IG failure modes (\%) among unverified interests, aggregated across all datasets. Categories are not mutually exclusive. \emph{Ins. Imp+}: insufficient implicit positive signals ($<$3); \emph{Ins. Exp+}: insufficient explicit positive signals ($<$2); \emph{Exc. Neg}: excessive negative signals. Insufficient positive evidence dominates across all models. Excessive negative contamination is bimodal.}
\label{tab:verification_failures}
\end{table}

Insufficient positive evidence dominates across all models. 92--99\% of failed interests lack enough explicit positive signals, and 74--97\% lack enough implicit positive signals. Excessive negative evidence is secondary and follows a bimodal pattern. Qwen2.5-72B and Qwen2-7B cite negatively-engaged content at $\sim$22\%, while DeepSeek-R1 and GPT-OSS-120B stay below 2\%. The cross-dataset pattern in Table~\ref{tab:merged_results} agrees. Signal type determines difficulty more than model identity, with Synthetic (rich multi-signal interactions) the easiest. This evidence-counting bottleneck matches results on structured-reasoning benchmarks beyond our setting. \citet{wu2024tablebench} similarly find that LLMs struggle with precise counting and aggregation over tabular evidence, which suggests the limitation is a general weakness in quantitative reasoning and not an artifact of our task.

\paragraph{Finding 3: Coverage Requires Scale and Reasoning; Specificity Does Not.}
The DeepSeek-R1 vs.\ R1-Distill-70B pair isolates the effect of model capacity within a single model lineage (same Arena IF Elo of 1395). The precision-recall decomposition reveals where distillation fails: R1-Distill-70B drops on \emph{both} IG$_P$ (54.9\% vs.\ 82.4\% on Synthetic) and IG$_R$ (28.8\% vs.\ 43.9\%), yielding substantially lower IG$_{F1}$ (37.3\% vs.\ 56.5\%). IS, however, remains nearly identical (78.0\% vs.\ 79.0\% on Synthetic), confirming that generating specific interest descriptions does not require the full reasoning capacity of the teacher model. The distilled model hallucinates more categories (lower IG$_P$) and discovers fewer genuine ones (lower IG$_R$). It also shows a fivefold increase in excessive negative evidence contamination (9.6\% vs.\ 1.8\% in Table~\ref{tab:verification_failures}). A control comparison sharpens this finding: Qwen2.5-72B, which has the same parameter count (72B) but no reasoning-specific training, achieves \emph{higher} IG$_{F1}$ than R1-Distill (45.5\% vs.\ 37.3\% on Synthetic). This suggests that reasoning-model distillation does not automatically transfer evidence-counting ability to smaller models, at least for the single distillation pair evaluated here. Generating descriptively specific interest labels (IS) does not require deep reasoning or large scale; accurately counting and attributing engagement signals (IG$_{F1}$) does. Figure~\ref{fig:ranking_ci} (Appendix~\ref{app:ranking_ci}) shows the 95\% confidence intervals for model rankings, confirming the statistical robustness of the gap between capable and small models. Figure~\ref{fig:ig_f1_distribution} (Appendix~\ref{app:score_dist}) illustrates the per-user score variance underlying these findings.

\paragraph{Finding 4: Context Length Improves Groundedness, Not Specificity.}
We investigate how UIH length affects model performance by varying the number of engagements per prompt for GPT-OSS-120B on the Synthetic dataset (Table~\ref{tab:uih_length}, Appendix~\ref{app:uih_ablation}). We select GPT-OSS-120B because it achieves the highest IG$_{F1}$ on the Synthetic dataset (Table~\ref{tab:merged_results}), providing the strongest baseline from which to isolate the effect of context length. All other experiments in this paper use the default UIH length of 100 engagements per prompt. IG$_R$ improves monotonically from 86.0\% to 96.8\% as UIH grows from 50 to 200 engagements, because more interaction history provides additional evidence objects that can meet the verification thresholds. IS increases modestly from 70.7\% to 74.6\% (+3.9pp), roughly half the rate of IG$_R$'s +10.8pp gain. Longer histories help the model accumulate more correctly-attributed backing evidence per category, but the effect on specificity is far smaller than on coverage. This confirms that the two metrics respond to different input axes at different magnitudes. The number of predicted interests decreases from 109 to 49: with more data, the model becomes more selective, producing fewer but better-grounded interests. This provides the cleanest evidence that IG and IS respond to different input axes, complementing the architectural evidence in Finding~3.

Together, these four findings characterize the capability profile required for LLM-based user understanding. A model needs format compliance to enter the game (F1), accurate evidence counting to ground predictions (F2), sufficient model capacity for multi-step signal attribution (F3), and context length to accumulate supporting evidence (F4). The precision-recall decomposition of IG shows that coverage (IG$_R$) is the universal bottleneck, while hallucination (IG$_P$) is not. All models achieve IG$_P >$ IG$_R$ (Table~\ref{tab:ig_pr_breakdown}), so predicted categories are mostly genuine but models discover only a fraction of verifiable interests. IG$_{F1}$ and IS are positively correlated overall (better models tend to score higher on both), but they respond differently to specific interventions in model architecture (Finding~3), context length (Finding~4), and dataset signal type (Finding~2). This validates their design as complementary evaluation dimensions. Prior LLM-as-a-Judge and xAI evaluation paradigms assess output plausibility or explanation quality, while our metrics directly verify predictions against behavioral evidence. This bridges the gap between the plausibility-focused evaluation common in related work and the verifiable grounding that user understanding demands.

\paragraph{Oracle Stability.}
The IG$_R$ oracle used for the reported scores is the union of verified categories across all evaluated models (Section~\ref{sec:ig}). To quantify how sensitive such a union oracle is to its constituent model set, we run a controlled ablation on the Synthetic dataset over the seven open-weight non-Llama models, which share a common serving configuration. For each of the 91K (user, interest category) pairs in this subset oracle, we count how many of the seven models independently verify it (Table~\ref{tab:oracle_stability}, Appendix~\ref{app:oracle_stability}). Only 1.3\% of these categories are verified by all seven, confirming that no single model, nor even the full subset, achieves complete interest coverage. Each model discovers categories that others miss: GPT-OSS-120B uniquely verifies 17.6\% of the subset oracle, while even Qwen2-7B contributes 0.6\% of categories found by no other model. At the same time, 36.8\% of categories are independently verified by three or more models, indicating substantial cross-model consensus on the core set of discoverable interests. 41.4\% are verified by only one model. These are disproportionately long-tail interests that individual models discover uniquely. The leave-one-out analysis (Table~\ref{tab:oracle_loo}, Appendix~\ref{app:oracle_stability}) shows that removing any single model reduces this subset oracle by at most 17.6\%. Models both agree on a common core and contribute unique discoveries. This validates the union construction as a reasonable upper bound, and confirms that IG$_R$ scores should not be compared across evaluation runs with different model sets.

\section{Conclusion}
\label{sec:conclusion}

We introduced GISTBench, a benchmark for evaluating LLM-based User Understanding in recommendation systems. Interest Groundedness (IG), decomposed into precision and recall components, and Interest Specificity (IS), restricted to verified categories, provide evaluation without requiring perfect ground-truth. These metrics correlate with user surveys ($\rho = 0.67$) and capture complementary behavioral dimensions. The precision-recall decomposition shows that the primary bottleneck across all evaluated models is interest category coverage, while hallucination is not. Our experiments show that current LLMs face challenges in evidence counting, implicit signal reasoning, and maintaining grounding capability at smaller model scales. These results point to concrete directions for future model development.

\paragraph{Future Work.}
Our benchmark uses text-only interaction histories. A natural next step is multimodal signals: image engagement, audio consumption, and video watch patterns all carry evidence about user interests that text metadata alone cannot capture. Another promising direction is integrating synthetic persona descriptions~\citep{ge2024scaling} into the user construction pipeline. Assigning each synthetic user cohort a natural language persona at Step~1 (Section~\ref{sec:dataset}) would enable evaluating whether models can reconcile explicit persona attributes (e.g., ``a college student interested in fitness and cooking'') with observed behavioral signals. This capability is important for conversational recommender systems that must maintain coherent user models across multi-turn interactions. On the temporal axis, our evaluation takes a single snapshot of each user, but interests drift: a user who binged cooking videos last month may have moved on. Detecting such shifts requires longitudinal evaluation that our current setup does not support. Finally, IG and IS could be used directly as reward signals for fine-tuning. Optimizing models to maximize verified coverage while minimizing hallucinated categories is a more targeted objective than general instruction following.

\paragraph{Limitations.}
Our benchmark assumes sufficient engagement data and does not address cold-start users with sparse histories. The verification thresholds, while validated by user surveys ($\rho = 0.67$), were designed for the current datasets and may require adjustment for different domains, cultures, or engagement taxonomies; Appendix~\ref{app:threshold} provides a threshold sensitivity analysis showing the relative model ranking is robust to $\pm1$ perturbations. The oracle used for IG Recall is constructed from the union of verified categories across all evaluated models, meaning that IG$_R$ scores are not directly comparable across evaluation runs with different model sets. Both the IG evidence-filtering judge and the IS retrieval judge use a single model (Llama-3.3-70B-Instruct). We provide inter-annotator agreement analysis in Appendix~\ref{app:human_eval}, but do not ablate on judge model choice. All evaluation is conducted in English on predominantly English-language content. Inference was performed on a single proprietary serving framework. Results for the open-weight models should be reproducible with equivalent generation parameters, but framework-level differences in tokenization or batching could introduce minor variation.

\paragraph{Acknowledgment.} This paper was written with the assistance of A.I. tools.

\bibliographystyle{assets/plainnat}
\bibliography{custom}

\beginappendix

\section{Interest Taxonomy Categories}
\label{app:taxonomy}

Tables~\ref{tab:l2-categories-1} and~\ref{tab:l2-categories-2} list all 325 interest categories from
the taxonomy used for interest category
normalization (Section~\ref{sec:normalization}). Categories are
listed alphabetically.

\begin{table}[p]
\caption{Interest categories used for taxonomy
normalization (A--L).}
\label{tab:l2-categories-1}
\centering
\footnotesize
\begin{multicols}{3}
\begin{enumerate}[leftmargin=*, itemsep=0pt, parsep=0pt, label=\arabic*.]
\raggedright
\item A.I. Art
\item ASMR
\item Accidents, Emergencies \& Tragedies
\item Aerospace Technology
\item Aircraft
\item Aliens \& UFO
\item Altruism \& Giving
\item American Football
\item Animal Comedy
\item Anime
\item Archaeology
\item Architecture
\item Art Exhibits \& Museums
\item Art History
\item Art Investment
\item Artificial Intelligence
\item Astrology \& Divination
\item Astronomy \& Space Exploration
\item Augmented Reality (AR) Filters
\item Automobile Collections
\item Automobile Reviews or Commentary
\item Automotive Lifestyle
\item Automotive Modifications
\item Automotive Purchases
\item Babies \& Toddlers
\item Backstage
\item Baked Goods \& Dessert
\item Ball Sports
\item Bands \& Musicians
\item Baseball
\item Basketball
\item Beachcombing \& Treasure-Seeking
\item Beauty Products
\item Behind the Scenes
\item Beverages
\item Birds
\item Birthdays
\item Board Games
\item Bugs \& Insects
\item Bushcraft
\item Business \& Financial News
\item Business Strategy
\item Buying a Home
\item Camping
\item Campus Comedy
\item Card \& Party Games
\item Cartoon / Animation
\item Celebrity Fandom
\item Celebrity Fashion
\item Celebrity Interviews
\item Celebrity Lifestyle
\item Celebrity Royals
\item Celebrity Spotlight
\item Children
\item Children's Books
\item Christmas
\item Cities \& Urban Exploration
\item Citizen Journalism
\item Classic Cars
\item Classroom Instruction
\item College Football
\item Combat Sports \& Martial Arts
\item Comedy Challenges
\item Comedy Dubbing
\item Comedy Remixes
\item Comics, Manga \& Graphic Novels
\item Competition / Reality Shows
\item Competitions \& Events
\item Computers \& Electronic Devices
\item Concerts
\item Cosplay \& LARP
\item Cover Songs
\item Cryptids
\item Cute Animals
\item Cute Cats
\item Cute Dogs
\item DIY Hacks, Tips \& Tricks
\item DIY Restorations \& Upscales
\item Daily Life
\item Dance Tutorials \& Instructions
\item Digital Commentary
\item Disability
\item Displayables, Collectibles \& Merch
\item Diving
\item Documentary
\item Drawing \& Painting
\item EMTs and Paramedics
\item Early Education
\item Ecology / Climatology
\item Electric Vehicles
\item Electronic Music
\item Entertainment News
\item Entrepreneurship \& Small Business
\item Epic Fails
\item Event Planning \& Venues
\item Exotic Pets
\item Extreme \& Adventure Sports
\item Fantasy \& Mythology
\item Farm Animals
\item Farms \& Ranches
\item Fashion, Dressing \& Styling Tips
\item Fast, Junk \& Snack Food
\item Festivals \& Traditions
\item Firefighting
\item Fishing
\item Floral
\item Food Cultivation
\item Food Demo \& Stills
\item Food During Travel
\item Food Recipes \& Cooking Tips
\item Food Review \& Critique
\item Formal Sciences
\item Furniture, Home Appliances \& Goods
\item Game Merchandise
\item Game Overview \& Strategy
\item Game Shows
\item Gameplay
\item Gaming Strategy
\item Gardening \& Planting
\item Geek Fandom
\item Get Ready with Me
\item Ghosts, Spirits \& Hauntings
\item Golf
\item Graphic Design
\item Gymnastics
\item Hair Care \& Styling
\item Halloween
\item Health \& Medical Sciences
\item Health Reports \& Updates
\item Healthy Foods
\item Healthy Living
\item Heavy Machinery
\item Hiking
\item History
\item Hobbies
\item Hockey
\item Holiday Decor
\item Home Dance
\item Home Renovations \& Repairs
\item Hot Takes
\item House \& Room Tours
\item Humorous Moments
\item Humorous Signs \& Images
\item Hunting
\item Inspiring Quotes \& Messages
\item Instrumental Music
\item Interior Design \& Decoration
\item Jewelry
\item Jokes \& Humorous Conversations
\item Kawaii
\item Kitchen \& Cooking Hacks
\item Kitchenware \& Tabletop
\item Knitting \& Sewing
\item Knowledge \& Trivia
\item Landscaping
\item Language Learning
\item Law Enforcement
\item Leasing \& Rental
\item Leather Goods
\item Lectures \& Presentations
\item Life Hacks
\item Lifestyle \& Urban Sports
\item Lip-Syncing
\item Literature \& Fiction
\item Live Streaming
\end{enumerate}
\end{multicols}
\end{table}

\begin{table}[p]
\caption{Interest categories used for taxonomy
normalization (L--Z), continued.}
\label{tab:l2-categories-2}
\centering
\small
\begin{multicols}{3}
\begin{enumerate}[leftmargin=*, itemsep=0pt, parsep=0pt, label=\arabic*., start=166]
\raggedright
\item Local, Regional or National Cuisine
\item Look at Me
\item Luxury \& Sports Cars
\item Luxury Homes
\item Machines \& Power Tools
\item Magazines \& Newspapers
\item Makeup Tutorials
\item Marine Life
\item Marriage \& Marital Life
\item Mechanical Engineering
\item Medical Practitioners
\item Memes
\item Mental \& Emotional Well-Being
\item Metalworking
\item Military \& Veteran
\item Mobile Games
\item Money \& Finance
\item Motorcycles
\item Motorsports \& Cycling
\item Movie \& TV Casts
\item Movie \& TV Characters
\item Movie \& TV Clips or Stills
\item Movies \& TV Reviews or Commentary
\item Mukbang
\item Muscle \& Strength Building
\item Music Reviews or Commentary
\item Music Tutorials \& Instructions
\item Music Videos
\item Musicals
\item Nail Art
\item Natural Sciences
\item Neurodiversity
\item Newborn \& Mommy
\item Non-Fiction Books
\item Nutrition \& Diet
\item Occult \& Mysticism
\item Online Courses
\item Opera
\item Organization, Storage \& Cleaning
\item Original Music
\item Other Fine Art
\item Other Sports
\item Outdoor Activities
\item Outdoor Equipment
\item Outfit \& Accessories
\item Outfit of the Day (OOTD)
\item Paper Crafts
\item Parent \& Child Relationship
\item Parent Life
\item Parenting Skills
\item Personal Growth
\item Pet Food \& Nutrition
\item Pet Health \& Wellness
\item Pet Rescue
\item Pet Stories
\item Pet Toys \& Accessories
\item Pet Training
\item Photography
\item Playground \& Yard
\item Podcasts
\item Poetry
\item Politics \& Government
\item Pranks
\item Precision Sports
\item Pregnancy
\item Pro Wrestling
\item Professional Development
\item Public Health
\item Puzzles \& Mysteries
\item Reaction Comedy
\item Real Estate
\item Recreational Vehicles
\item Religious Holidays
\item Restaurants \& Eateries
\item Road Trips
\item Role-Playing Games
\item Romantic Relationships \& Dating
\item Running
\item SAR (Search and Rescue)
\item Safety Cam Footage
\item Satires \& Spoofs
\item Scenic Vistas
\item Schools, Colleges \& Universities
\item Sci-Tech News
\item Science Experiments
\item Scripted Dramatic Content
\item Sculpture
\item Seasonal Trends
\item Secular Holidays
\item Shopping Guides
\item Short Films
\item Siblings \& Other Family
\item Sight-Seeing
\item Singer \& Dancer Stars
\item Sketches \& Skits
\item Skills Building
\item Skincare
\item Sleep, Meditation \& Mindfulness
\item Slime-Making
\item Small Apartments \& Tiny Homes
\item Smart \& Eco-Homes
\item Soap \& Candle Making
\item Soccer / Football
\item Social \& Friendship
\item Social Media Challenges
\item Social Science
\item Software \& Apps
\item Song Parodies
\item Special Occasions
\item Spiritual \& Religious
\item Sports Equipment
\item Sports News \& Commentary
\item Stand Up \& Roasting
\item Storytelling
\item Storytime
\item Street \& Public Art
\item Street Interviews \& Performers
\item Success Stories
\item TV Variety \& Talk Shows
\item Tattoos, Piercings \& Body Art
\item Teachers \& Teaching
\item Tech Builds \& Repairs
\item Teenagers
\item Tennis, Pickleball \& Racquetball
\item Theater \& Stage Plays
\item Theme Parks
\item Toys
\item Traditional \& Cultural Sports
\item Traditional \& Culture Dance
\item Traffic Reports \& Updates
\item Trains
\item Travel Planning \& Budgeting
\item Travel Tips \& Suggestions
\item Traveling with Kids
\item Trick Shots
\item Trucks
\item Unboxing
\item Unexplained Phenomena
\item Urban \& Hip Hop Dance
\item VR Gaming
\item Vegetarian \& Vegan
\item Veterinarian
\item Videography
\item Vlogs
\item Water Sports
\item Watercraft
\item Weather \& Natural Disasters
\item Web Series
\item Wedding Ceremonies \& Guests
\item Wedding Essentials
\item Weight Management
\item Wild Animals
\item Wilderness Exploration
\item Winter Sports
\item Woodworking
\item Workouts \& Training Regimens
\item Workplace Humor
\item Writers \& Writing
\item Yoga \& Pilates
\item eSports
\end{enumerate}
\end{multicols}
\end{table}

\section{Dataset Distribution Analysis}
\label{app:distributions}

\begin{figure}[t]
    \centering
    \begin{subfigure}[b]{0.48\columnwidth}
        \centering
        \includegraphics[width=\linewidth]{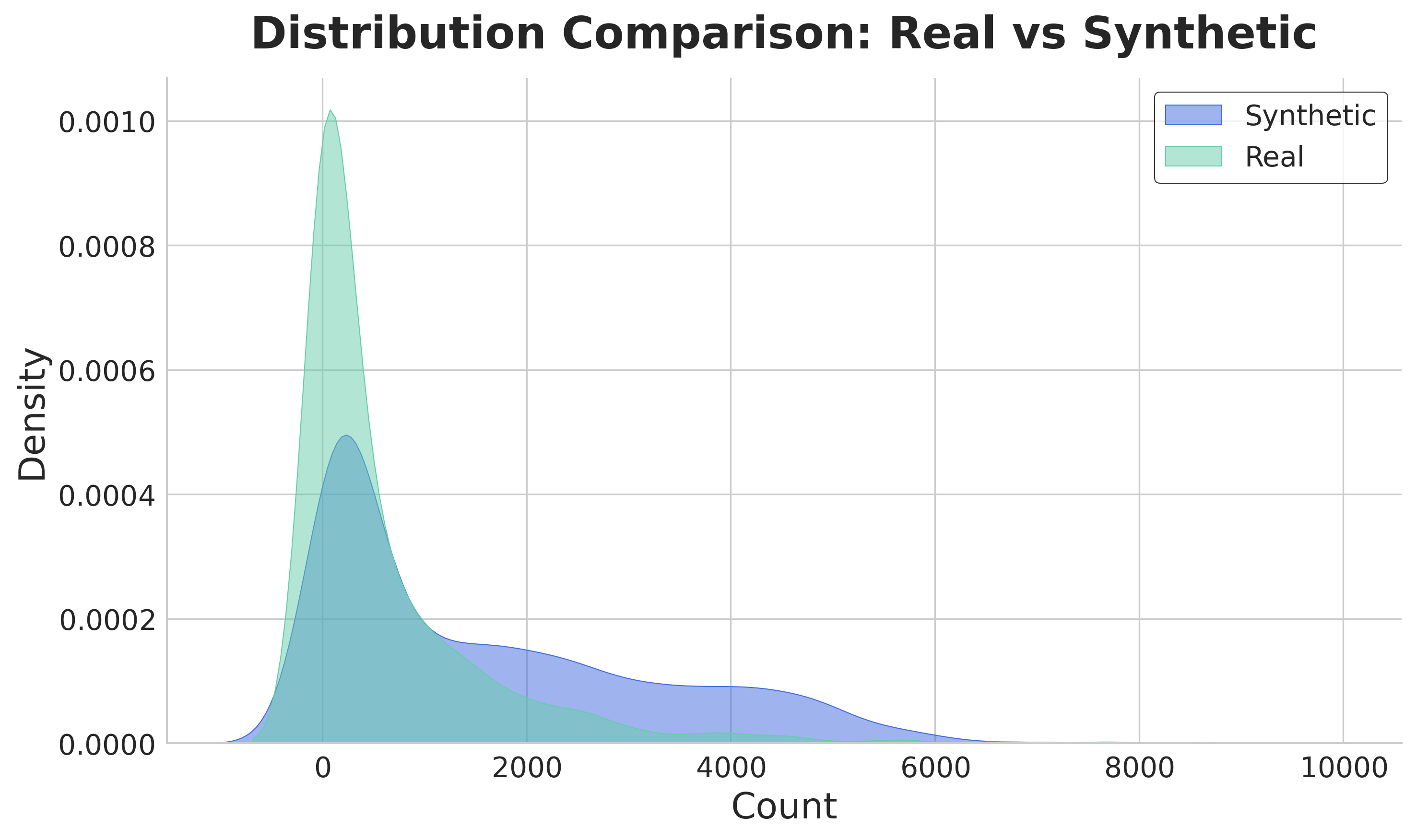}
        \caption{UIH Length Distribution}
        \label{fig:dist_kde}
    \end{subfigure}
    \hfill
    \begin{subfigure}[b]{0.48\columnwidth}
        \centering
        \includegraphics[width=\linewidth]{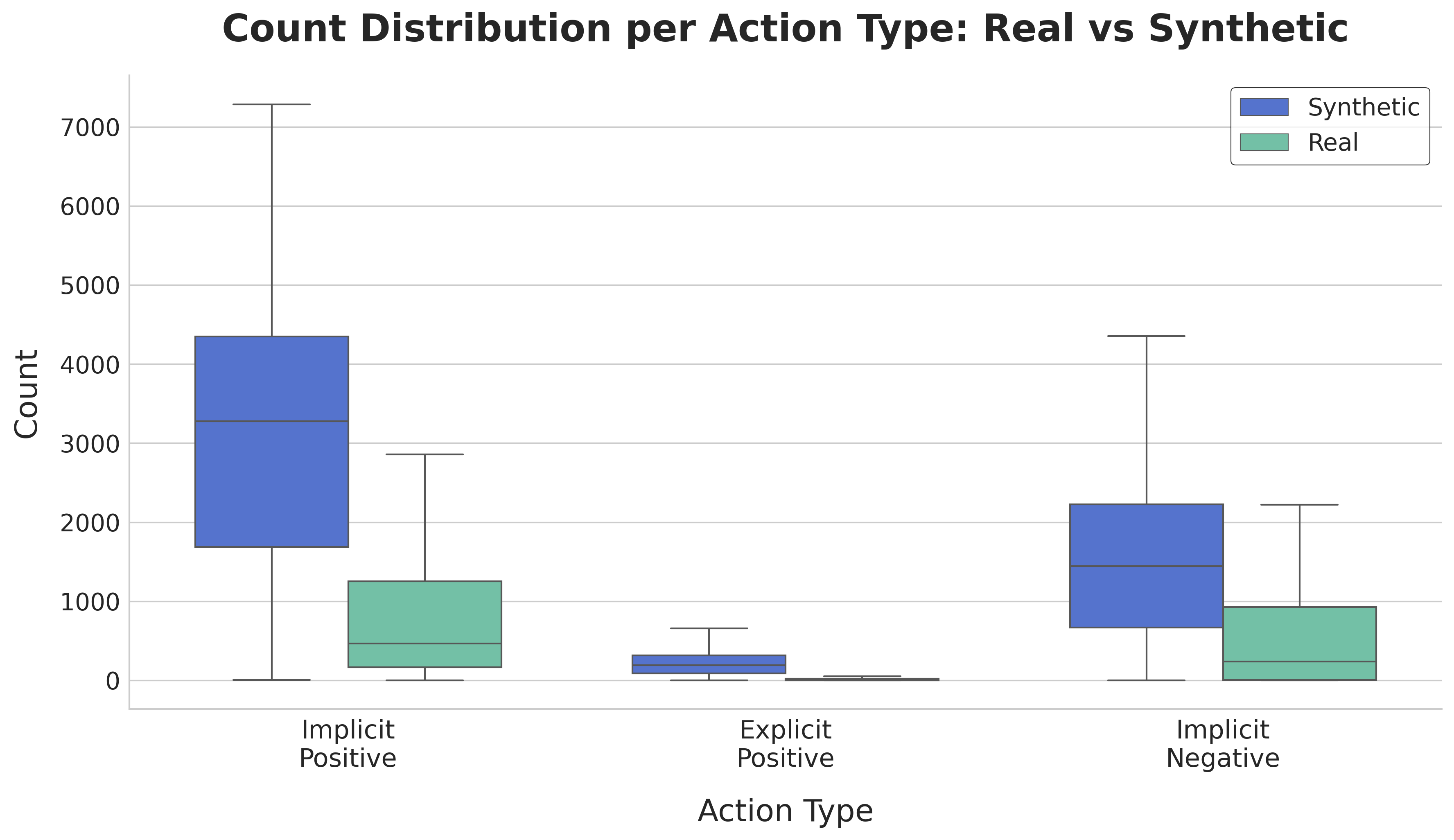}
        \caption{Action Type Distribution}
        \label{fig:dist_box}
    \end{subfigure}
    \caption{Comparison of synthetic and real UIH distributions. (a) Per-user engagement count densities show similar right-skewed patterns. (b) Box plots confirm that the relative ordering across action types is preserved.}
    \label{fig:distributions}
\end{figure}

Figure~\ref{fig:distributions} and Table~\ref{tab:distribution_stats} provide the full distributional comparison between synthetic and real UIHs referenced in Section~\ref{sec:dataset}.

\begin{table}[t]
\centering
\small
\setlength{\tabcolsep}{2pt}
\begin{tabular}{lcccc}
\toprule
\textbf{Action Type} & \textbf{t-stat} & \textbf{p-value} & \textbf{Syn. Mean} & \textbf{Real Mean} \\
\midrule
Implicit Negative & 15.65 & 4.80e-52 & 1569.39 & 796.30 \\
Implicit Positive & 21.44 & 6.02e-92 & 3308.83 & 1596.74 \\
Explicit Positive & 9.43 & 1.34e-20 & 105.66 & 43.79 \\
\bottomrule
\end{tabular}
\caption{Two-tailed t-tests comparing synthetic and real UIH engagement frequencies. Mean differences reflect intentional oversampling; relative proportions are preserved.}
\label{tab:distribution_stats}
\end{table}

\subsection{Amazon Music Sparsity}
\label{app:amazon_sparsity}

The Amazon Music dataset is extremely sparse, with 87.3\% of users having only 1 review. Under our verification framework, most users can verify at most one interest category, and only when that single review constitutes sufficient evidence (e.g., a 5-star explicit positive rating). We retain this dataset deliberately as a \emph{stress test} for our metrics under minimal-evidence conditions. With 1--2 reviews, very few interests can be verified, producing low IG$_{F1}$ scores that accurately reflect the information-theoretic limit of sparse behavioral data. This contrasts with denser datasets (Synthetic, KuaiRec) where models have abundant evidence to discover and verify interests. Amazon Music results measure benchmark behavior under sparsity. They do not provide a reliable ranking of model capability. Sample sizes are correspondingly smaller (89--117 users per model vs.\ 900--1000 for other datasets).

\section{Extended Analysis}
\label{app:extended}

This appendix presents additional visualizations that complement the tabular results in the main paper. Each figure shows a different perspective on the precision-recall tradeoff, model ranking significance, and per-user score distributions.

\subsection{IG Precision-Recall Tradeoff (Survey)}
\label{app:ig_pr_survey}

\begin{figure}[t]
\centering
\includegraphics[width=0.5\columnwidth]{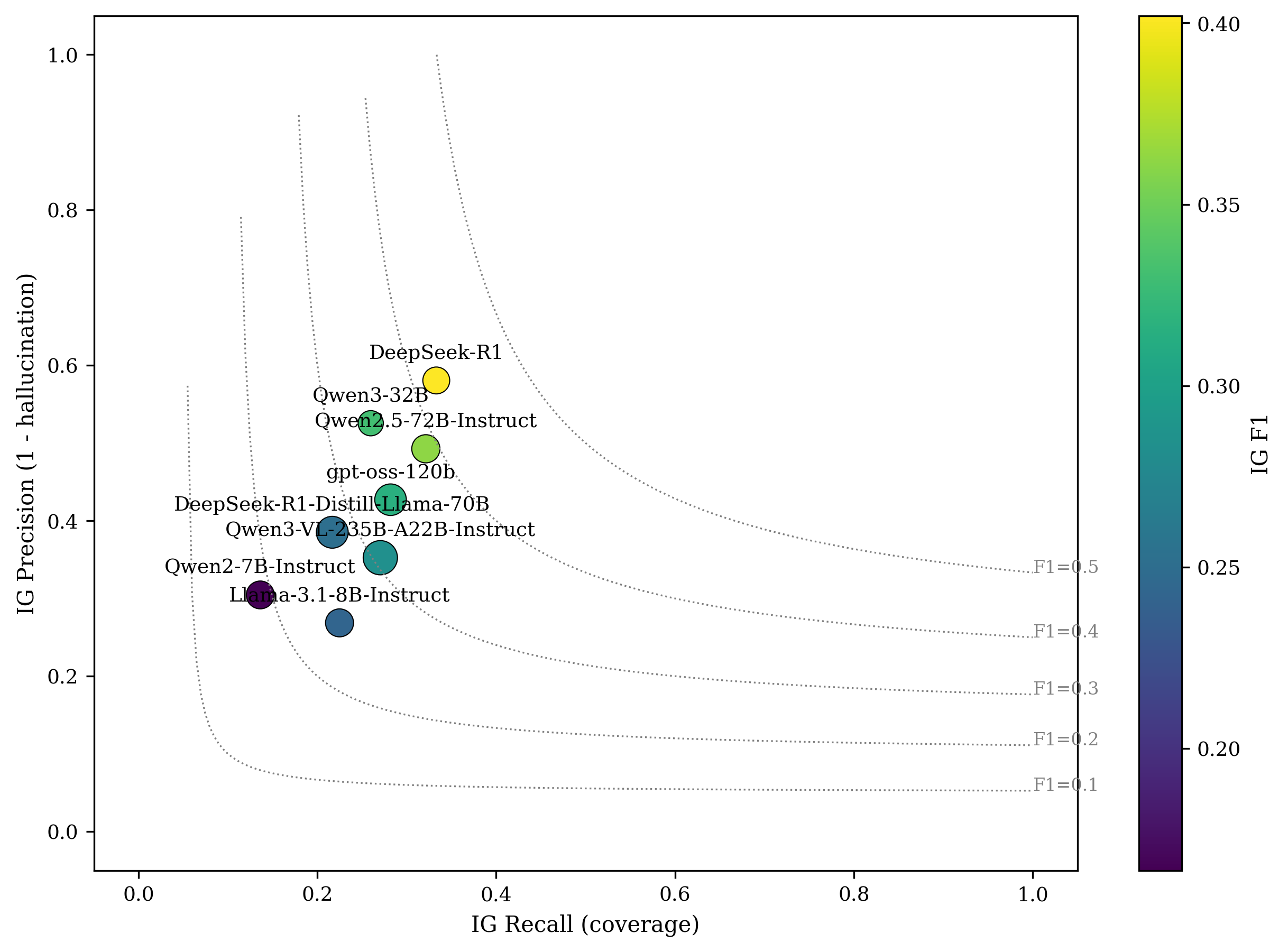}
\caption{IG Precision vs.\ IG Recall per model on the survey dataset. Bubble size encodes the average number of predicted interest categories; color encodes IG$_{F1}$. Dotted curves are iso-F1 contours at 0.1--0.5. All models cluster in the low-recall region (IG$_R < 0.4$), confirming that coverage is the universal bottleneck. DeepSeek-R1 achieves the highest precision (57.3\%) and IG$_{F1}$ (42.9\%). Qwen2.5-72B offers the best balance between precision and recall among capable models.}
\label{fig:ig_pr_survey}
\end{figure}

Figure~\ref{fig:ig_pr_survey} visualizes the IG precision-recall tradeoff on the survey dataset, complementing the tabular breakdown in Table~\ref{tab:survey_validation}. The iso-F1 contours reveal how far each model falls from balanced precision and recall: even the best model (DeepSeek-R1) sits well below the F1=0.5 contour. The bubble sizes show that models predicting more categories (larger bubbles) do not necessarily achieve higher IG$_{F1}$; Qwen3-VL-235B predicts many categories but achieves low IG$_P$ (25.0\%), indicating that quantity does not substitute for verification quality.

\subsection{Per-User Score Distributions (Benchmark)}
\label{app:per_user}

\begin{figure}[t]
\centering
\includegraphics[width=0.5\columnwidth]{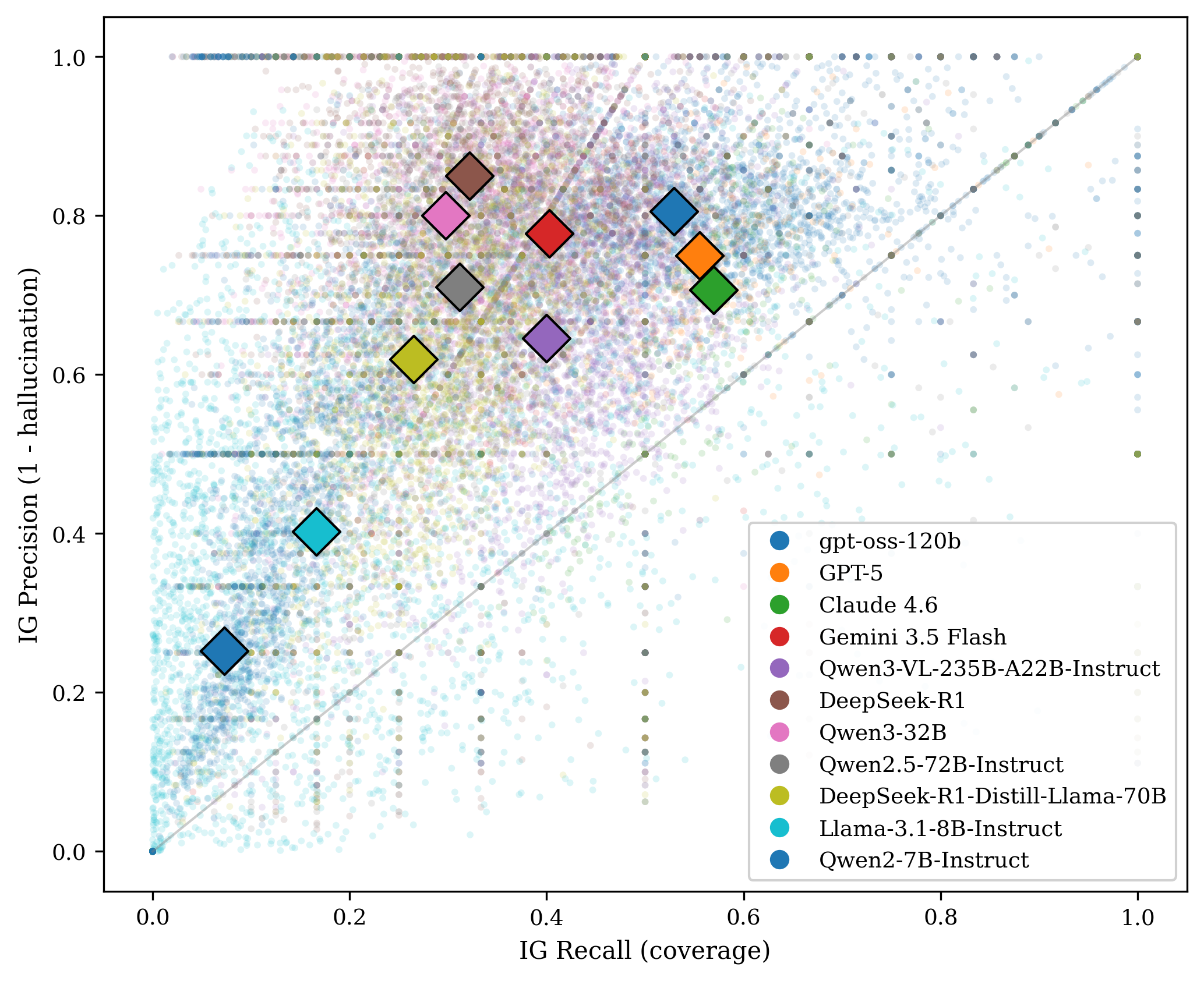}
\caption{Per-user IG Precision vs.\ IG Recall on the benchmark datasets, colored by model. Diamond markers show per-model medians; small dots show individual users ($\alpha=0.15$). The diagonal reference line ($y=x$) highlights the universal asymmetry IG$_P >$ IG$_R$. Within-model variance is substantial: even top models have many users with IG$_P < 0.5$. This reflects user-level difficulty variation.}
\label{fig:ig_pr_benchmark}
\end{figure}

Figure~\ref{fig:ig_pr_benchmark} reveals the per-user variance hidden behind median scores in Table~\ref{tab:merged_results}. Individual user scatter points show that within-model variance is substantial: even the best-performing models have many users with IG$_P < 0.5$. The median diamonds (large markers) show model-level tendencies. GPT-OSS-120B achieves the most balanced precision-recall profile, while DeepSeek-R1 clusters at high precision but low recall. The diagonal reference line confirms the universal asymmetry IG$_P >$ IG$_R$ at the individual user level, not just in aggregate. This motivates per-user analysis beyond median reporting.

\subsection{Model Ranking Significance}
\label{app:ranking_ci}

\begin{figure}[t]
\centering
\includegraphics[width=0.5\columnwidth]{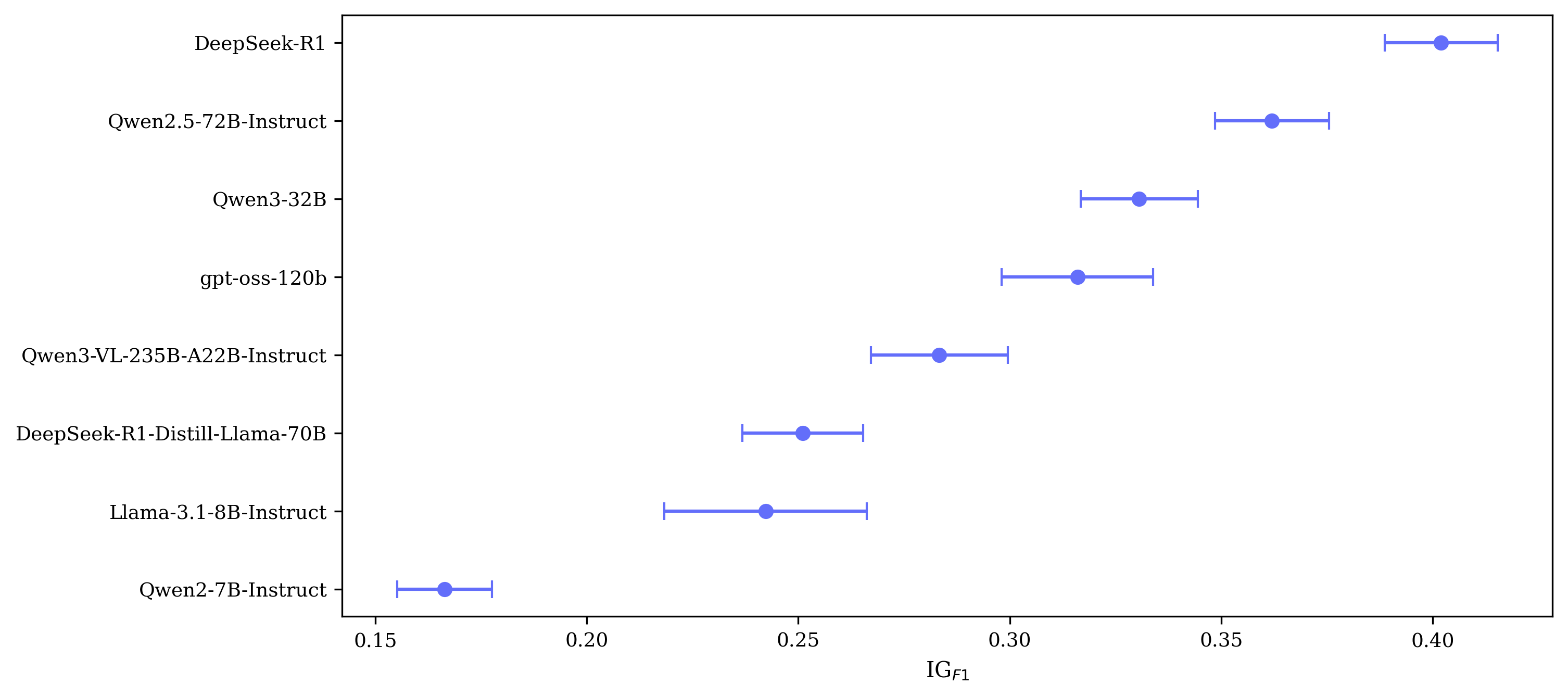}
\caption{Forest plot showing 95\% confidence intervals for mean IG$_{F1}$ scores on the survey dataset. Overlapping intervals indicate that ranking differences between adjacent models may not be statistically significant.}
\label{fig:ranking_ci}
\end{figure}

Figure~\ref{fig:ranking_ci} presents a forest plot of model rankings with 95\% confidence intervals for IG$_{F1}$ on the survey dataset. DeepSeek-R1 achieves the highest IG$_{F1}$ with a confidence interval clearly separated from all other models. The next three models (Qwen2.5-72B, Qwen3-32B, GPT-OSS-120B) have pairwise overlapping confidence intervals among adjacent ranks, suggesting that fine-grained ordering among them may not be statistically reliable. In contrast, clear separation exists between the capable model tier (IG$_{F1} > 0.25$) and the smaller models (Qwen2-7B and Llama-3.1-8B-Instruct), confirming that the capacity threshold identified in Finding~3 is statistically robust.

\subsection{Per-Model and Per-Dataset Taxonomy Coverage}
\label{app:taxonomy_coverage}

To assess whether individual models exhibit systematic blind spots in interest discovery, we analyze the coverage of the 325-category interest taxonomy (Section~\ref{sec:normalization}) at two granularities: per-model (aggregated across all datasets) and per-dataset (comparing single-model versus ensemble coverage).

\begin{table}[t]
\centering
\small
\setlength{\tabcolsep}{3pt}
\begin{tabular}{l|r|r|r}
\toprule
\textbf{Model} & \textbf{Unique Int.} & \textbf{Covered} & \textbf{Cov.\,\%} \\
\midrule
GPT-OSS-120B          & 127,495 & 325 & 100.0 \\
Qwen3-VL-235B         & 108,444 & 324 &  99.7 \\
DeepSeek-R1-Distill-70B & 37,340 & 323 &  99.4 \\
Llama-3.1-8B-Instruct   &  40,690 & 323 &  99.4 \\
DeepSeek-R1           &  78,534 & 320 &  98.5 \\
Qwen2-7B              &  21,999 & 318 &  97.8 \\
Qwen2.5-72B           &  22,177 & 318 &  97.8 \\
Qwen3-32B             &  39,002 & 317 &  97.5 \\
\bottomrule
\end{tabular}
\caption{Per-model taxonomy coverage aggregated across all datasets. \emph{Unique Int.}: number of distinct predicted interest strings mapped to the taxonomy. \emph{Covered}: number of the 325 taxonomy categories with $\geq$1 mapped interest. All eight models cover $\geq$97\% of categories.}
\label{tab:taxonomy_coverage}
\end{table}

\begin{figure}[htbp]
    \centering
    \includegraphics[width=0.5\columnwidth]{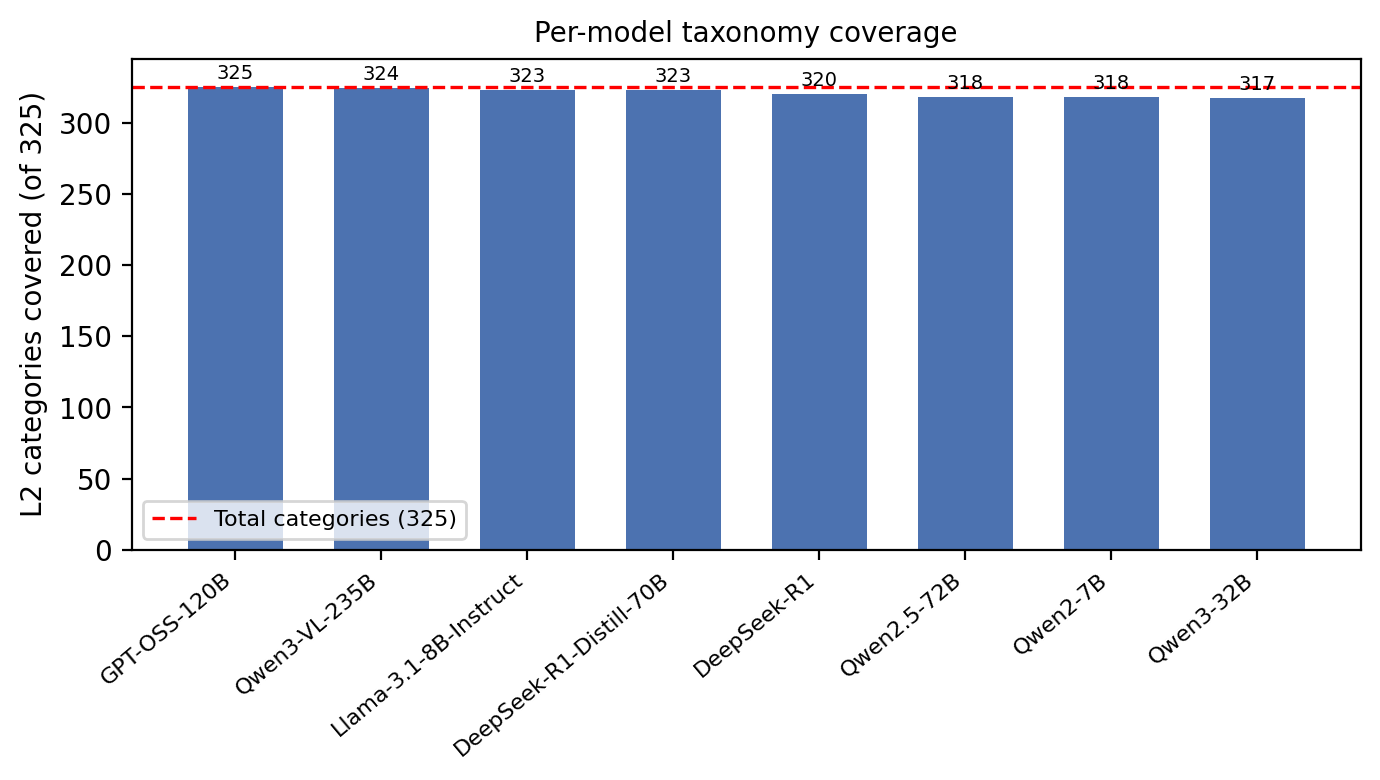}
    \caption{Taxonomy coverage by model, aggregated across all datasets. The red dashed line marks the total 325 taxonomy categories. All models cover $\geq$97.5\% of categories.}
    \label{fig:per_model_coverage_bar}
\end{figure}

Table~\ref{tab:taxonomy_coverage} and Figure~\ref{fig:per_model_coverage_bar} show that taxonomy coverage is near-universal across all models. Every model populates at least 317 of 325 categories ($\geq$97.5\%). GPT-OSS-120B achieves full coverage (325/325), while the remaining models miss only a handful of niche categories such as \emph{Playground \& Yard}, \emph{Art Investment}, and \emph{Augmented Reality (AR) Filters}. Coverage does not differentiate the models. Even the smallest models name interests spanning essentially the entire taxonomy, and the performance gaps in Table~\ref{tab:merged_results} stem from grounding and specificity, not breadth of interest vocabulary.

\begin{figure}[htbp]
    \centering
    \includegraphics[width=0.5\columnwidth]{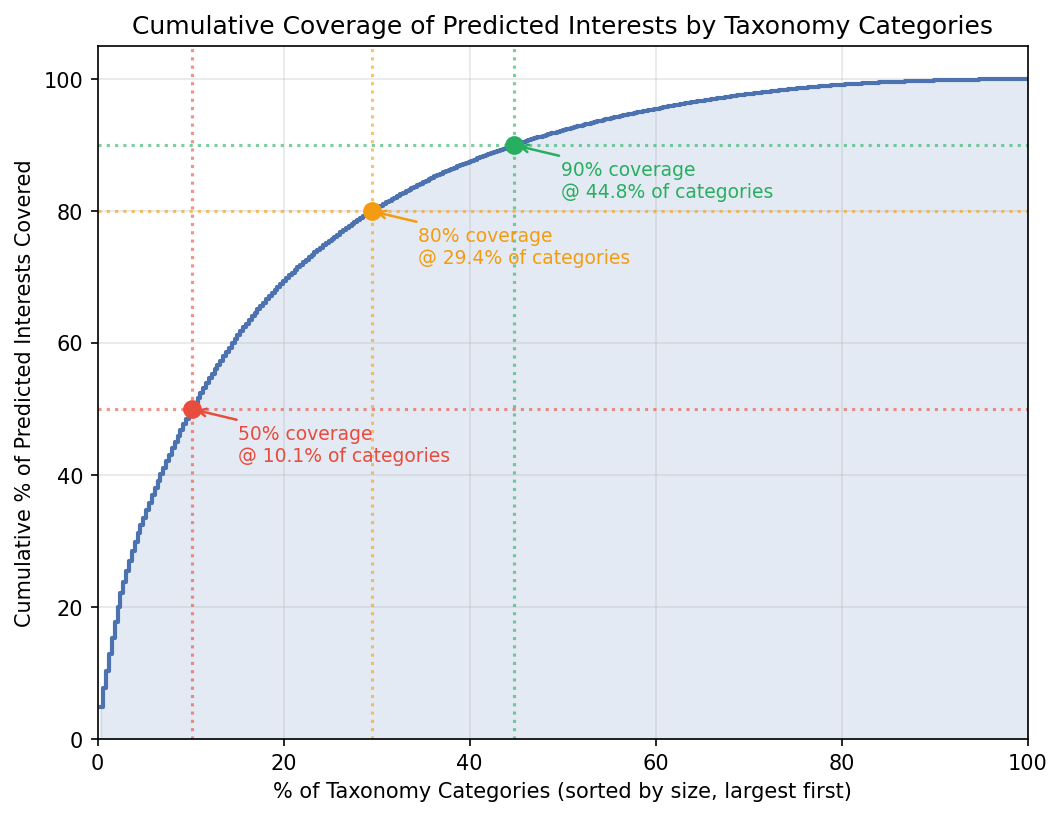}
    \caption{Cumulative distribution of predicted interests across taxonomy categories (sorted by size, largest first). The steep initial rise confirms heavy concentration: 10.1\% of categories capture 50\% of all predicted interests, and 29.4\% capture 80\%, reflecting the long-tailed nature of user engagement.}
    \label{fig:cumulative_coverage_cdf}
\end{figure}

Despite near-complete aggregate coverage, the distribution of interests across categories is heavily skewed for all models. Figure~\ref{fig:cumulative_coverage_cdf} visualizes this concentration: approximately 10\% of taxonomy categories account for 50\% of all mapped interests, 30\% of categories account for 80\%, and 45\% of categories account for 90\%. This long-tailed pattern (Gini coefficients range from 0.59 to 0.70 across models) mirrors the long-tailed nature of real user engagement distributions, where a small number of popular interest domains attract the bulk of user activity.

\begin{figure}[htbp]
    \centering
    \begin{subfigure}[b]{0.48\columnwidth}
        \centering
        \includegraphics[width=\linewidth]{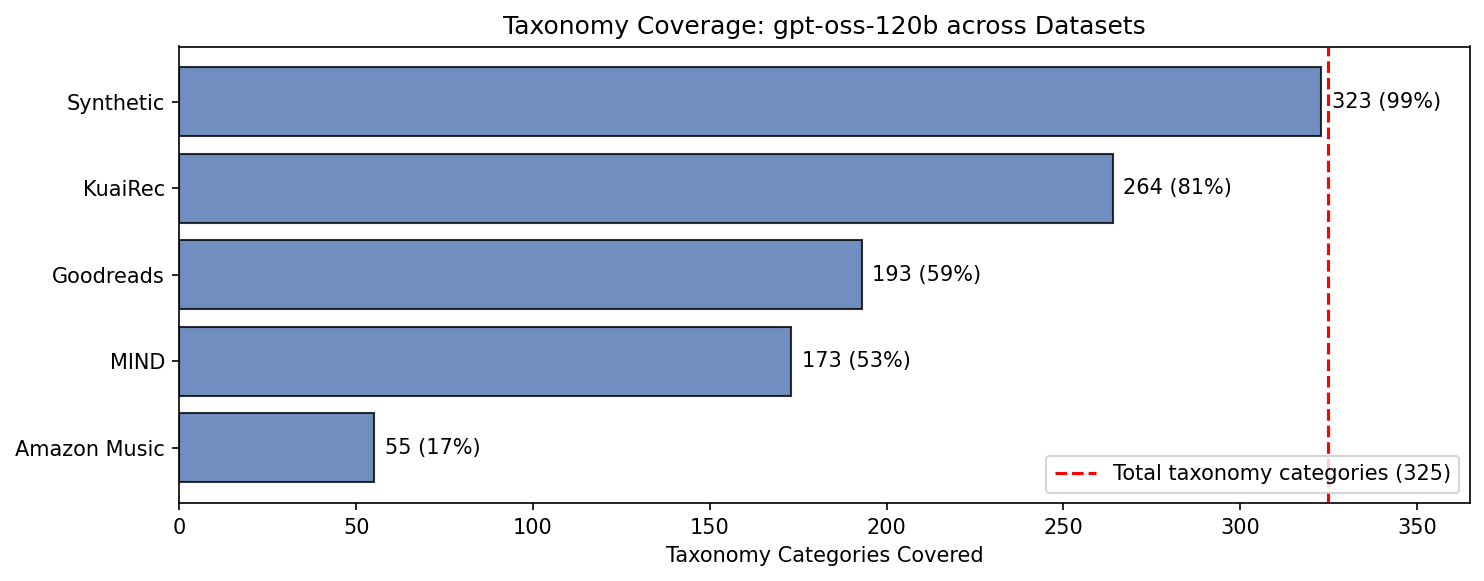}
        \caption{GPT-OSS-120B coverage per dataset.}
        \label{fig:gpt_coverage}
    \end{subfigure}
    \hfill
    \begin{subfigure}[b]{0.48\columnwidth}
        \centering
        \includegraphics[width=\linewidth]{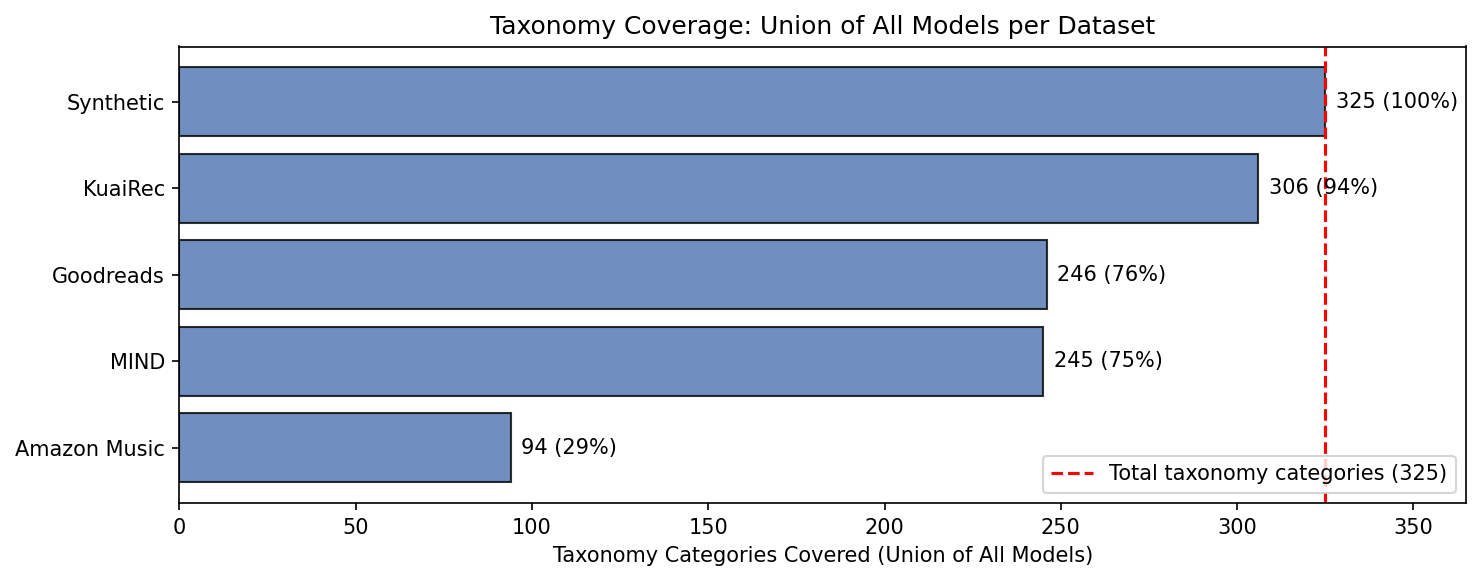}
        \caption{Union of all models per dataset.}
        \label{fig:union_coverage}
    \end{subfigure}

    \caption{Taxonomy coverage across datasets for (a) GPT-OSS-120B alone and (b) the union of all evaluated models. The red dashed line marks the total 325 taxonomy categories. Ensembling substantially improves coverage on mid-range datasets (e.g., MIND: 53\% $\to$ 75\%; Goodreads: 59\% $\to$ 76\%) but has limited effect where a single model already saturates (Synthetic) or where data sparsity fundamentally constrains discovery (Amazon Music).}
    \label{fig:taxonomy_coverage_datasets}
\end{figure}

\paragraph{Per-Dataset Coverage.}
Coverage varies substantially across datasets. For the best-performing single model (GPT-OSS-120B), coverage ranges from 323/325 (99\%) on the Synthetic dataset down to 55/325 (17\%) on Amazon Music (Figure~\ref{fig:gpt_coverage}). This variation is driven by dataset characteristics: the Synthetic dataset provides rich multi-signal interactions across diverse content, enabling broad interest discovery, while Amazon Music's extreme sparsity (87.3\% of users have only 1 review) severely limits the range of discoverable interest categories.

Comparing single-model coverage (Figure~\ref{fig:gpt_coverage}) with the union across all models (Figure~\ref{fig:union_coverage}) quantifies the benefit of model ensembling. On the Synthetic dataset, GPT-OSS-120B already covers 99\% of categories, leaving negligible room for improvement (union: 100\%). On MIND and Goodreads, however, the ensemble lifts coverage by 22 and 17 percentage points respectively (MIND: 173 $\to$ 245; Goodreads: 193 $\to$ 246), indicating that different models discover complementary interest categories in these domains. Amazon Music sees a more modest absolute gain (55 $\to$ 94), reflecting the fundamental constraint of data sparsity rather than model capability.

Taxonomy coverage is not a differentiating axis among capable models when aggregated across datasets. The bottleneck identified in Finding~2 (evidence counting) and Finding~3 (coverage requiring scale) operates \emph{within} categories; it is not a failure to span \emph{across} them. Models fail because they cannot accumulate sufficient verified evidence within the domains they do discover, not because they miss entire interest domains. The per-dataset analysis further shows that dataset signal richness is the primary determinant of how many interest categories can be populated, more than model identity is.

\subsection{Score Distribution Analysis}
\label{app:score_dist}

\begin{figure}[t]
\centering
\includegraphics[width=0.5\columnwidth]{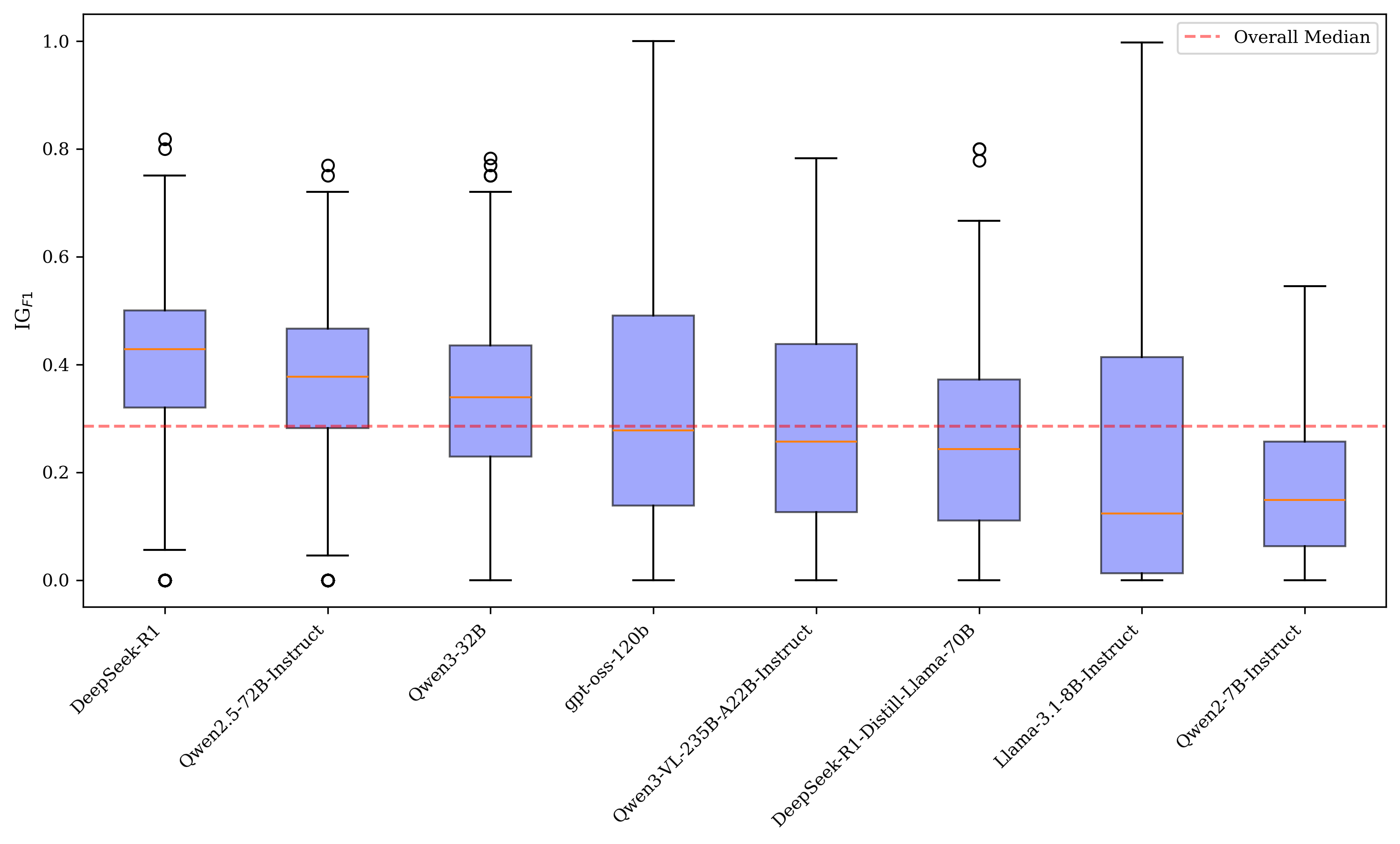}
\caption{Box plot of per-user IG$_{F1}$ scores by model on the survey dataset. The red dashed line shows the overall median. Even top models have users at IG$_{F1} = 0$, showing substantial within-model variance.}
\label{fig:ig_f1_distribution}
\end{figure}

Figure~\ref{fig:ig_f1_distribution} shows the full distribution of per-user IG$_{F1}$ scores. The interquartile ranges reveal that model performance varies substantially across users. DeepSeek-R1's median of ${\sim}0.43$ masks a distribution spanning from 0.0 to over 0.8. This spread comes from user-level difficulty variation, since users with sparse or ambiguous interaction histories are inherently harder to model. Qwen2-7B and Llama-3.1-8B-Instruct have the lowest medians, consistent with their bottom-tier IG$_{F1}$ (Finding~1).

\section{Qualitative Walkthrough}
\label{app:walkthrough}

To illustrate how our metrics operate in practice, we trace two predicted interests through the full evaluation pipeline for a single synthetic user. Table~\ref{tab:walkthrough_evidence} shows the cited evidence objects with their content and engagement type, and Table~\ref{tab:walkthrough_verdict} summarizes the IG verification and IS retrieval outcomes.

\begin{table}[t]
\centering
\small
\resizebox{\columnwidth}{!}{
\begin{tabular}{llll}
\toprule
\textbf{Interest} & \textbf{Object} & \textbf{Content Description} & \textbf{Engagement} \\
\midrule
\multirow{4}{*}{\makecell[l]{``NBA Basketball\\Highlights''}}
  & vid\_12 & \#NBA \#LeBron ``LeBron's game-winning dunk vs Celtics'' & like (\textbf{exp+}) \\
  & vid\_34 & \#Basketball ``Top 10 plays of the week'' & watch 90s (\textbf{imp+}) \\
  & vid\_56 & \#NBAPlayoffs ``Conference finals recap'' & watch 120s (\textbf{imp+}) \\
  & vid\_78 & \#NBA \#GameDay ``Buzzer beater compilation 2024'' & share (\textbf{exp+}) \\
\midrule
\multirow{4}{*}{\makecell[l]{``Italian Cooking\\Recipes''}}
  & vid\_91 & \#ItalianFood ``How to make carbonara at home'' & watch 45s (\textbf{imp+}) \\
  & vid\_23 & \#Cooking ``Knife skills tutorial for beginners'' & skip (\textbf{imp$-$}) \\
  & vid\_45 & \#FoodTok ``Quick weeknight dinner ideas'' & skip (\textbf{imp$-$}) \\
  & vid\_67 & \#ItalianCuisine ``Making fresh ravioli from scratch'' & watch 30s (\textbf{imp+}) \\
\bottomrule
\end{tabular}
}
\caption{Evidence objects cited by the model for two predicted interests. Each object includes the content description (hashtags and title) and the user's engagement type, classified as implicit positive (\textbf{imp+}: watch $\geq$30s), explicit positive (\textbf{exp+}: like, share), or implicit negative (\textbf{imp$-$}: skip).}
\label{tab:walkthrough_evidence}
\end{table}

\begin{table}[t]
\centering
\small
\setlength{\tabcolsep}{3pt}
\begin{tabular}{l|cccc|c|c}
\toprule
\textbf{Interest} & \textbf{Imp+} & \textbf{Exp+} & \textbf{Imp$-$} & \textbf{Exp$-$} & \textbf{IG Verdict} & \textbf{IS} \\
\midrule
NBA Basketball & 2 & 2 & 0 & 0 & \ding{51} Verified & 3/4 \\
Italian Cooking & 2 & 0 & 2 & 0 & \ding{55} Failed & --- \\
\bottomrule
\end{tabular}
\caption{IG verification and IS retrieval results. ``NBA Basketball Highlights'' passes verification ($\geq$2 explicit positives) and achieves IS = 3/4 (the judge correctly identifies 3 of 4 backing objects from a pool of 50). ``Italian Cooking Recipes'' fails: despite 2 implicit positives, it has 0 explicit positives and 2 implicit negatives (skips), indicating insufficient evidence of genuine interest. IS is not computed for unverified interests.}
\label{tab:walkthrough_verdict}
\end{table}

\paragraph{IG Verification.} For ``NBA Basketball Highlights,'' the model cited 4 objects, of which 2 are explicit positives (like, share) and 2 are implicit positives (extended watches). This exceeds the verification threshold ($\geq$2 explicit positives), so the interest is \emph{verified}. For ``Italian Cooking Recipes,'' while the user watched two cooking videos for $\geq$30s (implicit positives), they also skipped two others (implicit negatives), and no explicit positive signals (likes, shares) are present. With 0 explicit positives and 2 implicit negatives approaching the contamination threshold ($\leq$3 implicit negatives), this interest \emph{fails} verification; the engagement evidence is ambiguous rather than confirmatory.

\paragraph{IS Retrieval.} For the verified interest ``NBA Basketball Highlights,'' the IS judge receives the interest text and a pool of 50 candidate objects (4 true evidence + 46 distractors drawn from other users). The judge correctly identifies 3 of 4 backing objects, yielding IS = 0.75 for this interest. IS is not computed for ``Italian Cooking Recipes'' because it failed IG verification, which prevents ungrounded interests from inflating specificity scores.

\section{UIH Length Ablation}
\label{app:uih_ablation}

Table~\ref{tab:uih_length} reports the effect of varying User Interaction History (UIH) length on GPT-OSS-120B performance using the Synthetic dataset. We vary the number of engagements per prompt from 50 to 200 while holding all other parameters constant. As UIH length increases, both IG$_R$ and IS improve, but at markedly different rates: IG$_R$ gains +10.8 percentage points while IS gains only +3.9pp, confirming that the two metrics respond to different input axes. The number of predicted interests decreases monotonically from 109 to 49, indicating that with more context the model becomes more selective, consolidating related signals into fewer, better-grounded interests rather than producing additional predictions.

\begin{table}[t]
\centering
\small
\begin{tabular}{l|c|c|c|c}
\toprule
\textbf{UIH Length} & \textbf{50} & \textbf{100} & \textbf{150} & \textbf{200} \\
\midrule
IG$_P$ (\%) & 74.4 & 80.9 & 84.9 & 88.9 \\
IG$_R$ (\%) & 86.0 & 91.0 & 93.8 & 96.8 \\
IG$_{F1}$ (\%) & 79.7 & 85.6 & 89.2 & 92.5 \\
IS (\%) & 70.7 & 72.4 & 73.9 & 74.6 \\
\# Interests & 109 & 92 & 72 & 49 \\
\bottomrule
\end{tabular}
\caption{Effect of UIH length on GPT-OSS-120B performance (Synthetic dataset). IG$_R$ and IS both improve with more context, but at markedly different rates (+10.8pp vs.\ +3.9pp), confirming the two metrics respond to different input axes.}
\label{tab:uih_length}
\end{table}

\section{Oracle Stability}
\label{app:oracle_stability}

Table~\ref{tab:oracle_stability} reports the distribution of oracle category discovery across models on the Synthetic dataset. Table~\ref{tab:oracle_loo} shows the leave-one-out impact: how many oracle categories each model uniquely provides.

\begin{table}[t]
\centering
\small
\begin{tabular}{c|r|r}
\toprule
\textbf{Models verifying} & \textbf{Count} & \textbf{\%} \\
\midrule
1 & 37,672 & 41.4 \\
2 & 19,929 & 21.9 \\
3 & 13,325 & 14.6 \\
4 & 9,680 & 10.6 \\
5 & 6,011 & 6.6 \\
6 & 3,276 & 3.6 \\
7 & 1,206 & 1.3 \\
\midrule
\textbf{Total} & \textbf{91,099} & \textbf{100} \\
\bottomrule
\end{tabular}
\caption{Oracle category discovery distribution on the Synthetic dataset (7 models). Each row counts how many (user, interest category) pairs in the oracle are independently verified by exactly $k$ models. 36.8\% of oracle entries are verified by 3+ models (cross-model consensus), while 41.4\% depend on a single model.}
\label{tab:oracle_stability}
\end{table}

\begin{table}[t]
\centering
\small
\begin{tabular}{l|r|r|r}
\toprule
\textbf{Model} & \textbf{Verified} & \textbf{Sole provider} & \textbf{\% oracle} \\
\midrule
GPT-OSS-120B & 60,971 & 16,065 & 17.6 \\
Qwen3-VL-235B & 48,260 & 8,844 & 9.7 \\
Qwen2.5-72B & 36,778 & 5,550 & 6.1 \\
DeepSeek-R1 & 26,158 & 3,569 & 3.9 \\
R1-Distill-70B & 19,696 & 1,774 & 1.9 \\
Qwen3-32B & 13,900 & 1,308 & 1.4 \\
Qwen2-7B & 8,615 & 562 & 0.6 \\
\bottomrule
\end{tabular}
\caption{Leave-one-out oracle impact on the Synthetic dataset. ``Verified'' = total (user, interest category) pairs this model verifies. ``Sole provider'' = pairs verified \emph{only} by this model (lost if removed). GPT-OSS-120B (the broadest-coverage model) uniquely provides 17.6\% of the oracle, while every model contributes at least some unique categories.}
\label{tab:oracle_loo}
\end{table}

\section{Human Annotation Study for Judge Validation}
\label{app:human_eval}

We conduct a human annotation study to validate the two LLM judges used in our pipeline. Four annotators independently labeled 1{,}000 items for each task (2{,}000 annotations per annotator).

\paragraph{Annotation Tasks.}
We evaluate the two judgment operations used by our metrics:

\begin{itemize}[leftmargin=1.5em]
    \item \textbf{Evidence Relevance (IG):} Given a predicted interest, an engagement type, and a single content item, the annotator labels whether the item is \emph{truly relevant} to the interest (Yes/No). This corresponds to the per-item filtering step in IG verification (Section~\ref{sec:ig}).
    \item \textbf{Interest--Item Association (IS):} Given a predicted interest and a single content item description (no engagement context), the annotator labels whether the item supports inferring this interest (True/False). This corresponds to the per-item retrieval decision in IS evaluation (Section~\ref{sec:is}).
\end{itemize}

Annotators receive one (interest, item) pair at a time and produce a binary label. The 1{,}000 items per task were sampled from the Synthetic dataset evaluation outputs, stratified across users, models, and interest categories.

\paragraph{Inter-Annotator Agreement.}
Table~\ref{tab:human_kappa} reports pairwise Cohen's $\kappa$ for all annotator pairs.

\begin{table}[t]
\centering
\small
\setlength{\tabcolsep}{4pt}
\begin{tabular}{l|cccc}
\toprule
\multicolumn{5}{c}{\textbf{Evidence Relevance (IG)}} \\
\midrule
 & A$_1$ & A$_2$ & A$_3$ & A$_4$ \\
\midrule
A$_1$ & --- & 0.40 & \textbf{0.70} & 0.47 \\
A$_2$ &     & ---  & 0.44 & 0.43 \\
A$_3$ &     &      & ---  & 0.65 \\
A$_4$ &     &      &      & ---  \\
\midrule
\multicolumn{5}{c}{\textbf{Interest--Item Association (IS)}} \\
\midrule
 & A$_1$ & A$_2$ & A$_3$ & A$_4$ \\
\midrule
A$_1$ & --- & \textbf{0.88} & 0.83 & 0.81 \\
A$_2$ &     & ---  & 0.86 & 0.84 \\
A$_3$ &     &      & ---  & \textbf{0.94} \\
A$_4$ &     &      &      & ---  \\
\bottomrule
\end{tabular}
\caption{Pairwise Cohen's $\kappa$ between four annotators ($N = 1{,}000$ per task). IG average pairwise $\kappa = 0.52$ (moderate); IS average pairwise $\kappa = 0.86$ (almost perfect).}
\label{tab:human_kappa}
\end{table}

IS agreement is high: all six pairs reach $\kappa \geq 0.81$ (average 0.86, almost perfect~\cite{landis1977measurement}). Deciding whether a content item supports a given interest is a relatively objective judgment. IG agreement is lower and more variable ($\kappa$ from 0.40 to 0.70, average 0.52). The variation is driven by differing positive-label rates: A$_2$ and A$_4$ label 87\% and 90\% of items as relevant, while A$_1$ labels 79\%. This gap in strictness lowers pairwise $\kappa$ even when raw agreement remains above 85\%.

\paragraph{Majority-Vote Agreement.}
Table~\ref{tab:human_majority} reports each annotator against the majority vote ($\geq$3 of 4 annotators).

\begin{table}[t]
\centering
\small
\setlength{\tabcolsep}{4pt}
\begin{tabular}{l|ccc|ccc}
\toprule
& \multicolumn{3}{c|}{\textbf{IG}} & \multicolumn{3}{c}{\textbf{IS}} \\
 & Agree\% & $\kappa$ & F$_1$ & Agree\% & $\kappa$ & F$_1$ \\
\midrule
A$_1$ & 93.7 & 0.81 & 0.96 & 98.4 & 0.93 & 0.94 \\
A$_2$ & 88.0 & 0.56 & 0.93 & 99.1 & 0.96 & 0.96 \\
A$_3$ & 96.7 & \textbf{0.89} & \textbf{0.98} & 97.6 & 0.90 & 0.91 \\
A$_4$ & 90.8 & 0.65 & 0.95 & 97.1 & 0.88 & 0.90 \\
\bottomrule
\end{tabular}
\caption{Agreement with majority vote ($\geq$3/4). All annotators reach $\geq$88\% agreement on IG and $\geq$97\% on IS.}
\label{tab:human_majority}
\end{table}

On IG, A$_3$ best matches the group consensus ($\kappa = 0.89$, F$_1 = 0.98$). A$_2$ has lower $\kappa$ (0.56) because of its more inclusive labeling, but still agrees on 88\% of items. On IS, all four annotators reach $\kappa \geq 0.88$.

\paragraph{Unanimity Analysis.}
Table~\ref{tab:human_unanimity} shows how often annotators fully agree.

\begin{table}[t]
\centering
\small
\begin{tabular}{l|cc}
\toprule
\textbf{Agreement level} & \textbf{IG} & \textbf{IS} \\
\midrule
Unanimous (4/4 or 0/4) & 77.0\% & 94.0\% \\
3--1 split              & 15.2\% &  4.2\% \\
2--2 split              &  7.8\% &  1.8\% \\
\bottomrule
\end{tabular}
\caption{Agreement-level distribution. 94\% of IS items and 77\% of IG items receive unanimous labels.}
\label{tab:human_unanimity}
\end{table}

94\% of IS items receive a unanimous label, and only 1.8\% produce a 2--2 split. IG unanimity is 77\%, with 7.8\% of items producing a 2--2 split. These borderline IG items tend to involve tangentially related content (e.g., ``Self-driving car technology'' labeled under ``Electric Vehicle Technology''). The 3--1 splits (15.2\%) reflect one annotator applying a stricter or looser threshold than the other three.

\paragraph{Implications.}
IS agreement ($\bar{\kappa} = 0.86$) confirms that the IS retrieval task is well-defined and produces reliable labels. IG agreement ($\bar{\kappa} = 0.52$) is moderate, but the best-aligned annotator reaches $\kappa = 0.89$ against the majority, showing that majority-aggregated labels are a reliable reference. These results support the use of automated judges for both the IG evidence-filtering and IS retrieval stages of the GISTBench pipeline.

\section{MIND Dataset Analysis}
\label{app:mind}

Among the public datasets we evaluate, MIND~\cite{wu2020mind} is uniquely ill-suited to evidence-based interest verification, and we therefore exclude it from the main results (Table~\ref{tab:merged_results}). This appendix explains why.

\paragraph{Impoverished Signal Taxonomy.}
MIND exposes only a single positive signal (\emph{click}) and a single negative signal (\emph{non-click impression}). We map clicks to explicit positive and non-clicks to implicit negative; MIND has no implicit-positive (e.g.\ dwell-time) signal. Its engagement distribution is also extremely skewed: across the sampled users, roughly $3.5\%$ of impressions are clicks and $96.5\%$ are non-clicks. Every other dataset in our benchmark provides at least three graded engagement levels (Table~\ref{tab:dataset_comparison}); MIND collapses to a near-binary click signal, which both starves the verification predicate of evidence and motivates the rich multi-signal design of our synthetic dataset.

\paragraph{Per-Dataset Predicate.}
The only positive-evidence clause of the general predicate (Section~\ref{sec:ig}) that can fire on MIND is the explicit-only one, so the MIND predicate reduces to
\begin{equation}
\label{eq:phi_mind}
\phi_{\text{MIND}} = \mathbf{1}\bigl[\,n^+_{\text{exp}} \geq 2 \;\wedge\; n^-_{\text{imp}} \leq 3\,\bigr],
\end{equation}
i.e.\ an interest is verified iff at least two of its judge-approved citations are clicks and at most three are non-clicks. The implicit-positive and hybrid clauses never apply.

\section{Threshold Sensitivity Analysis}
\label{app:threshold}

The verification predicate (Section~\ref{sec:ig}) uses fixed evidence thresholds ($\geq$2 explicit positives, $\geq$3 implicit positives, $\leq$3 implicit negatives). To test whether the benchmark's conclusions depend on these specific values, we re-score the Synthetic dataset under three predicate configurations that shift every threshold by one step (Table~\ref{tab:threshold_grid}), without any additional LLM inference. Only the verification predicate changes.

\begin{table}[t]
\centering
\small
\setlength{\tabcolsep}{4pt}
\begin{tabular}{l|cccc}
\toprule
\textbf{Setting} & \textbf{Exp+} & \textbf{Imp+} & \textbf{Hybrid} & \textbf{Imp$-$ cap} \\
\midrule
Lenient ($-1$) & $\geq$1 & $\geq$1 & --- & $\leq$4 \\
Base (paper)   & $\geq$2 & $\geq$3 & $\geq$1 \& $\geq$2 & $\leq$3 \\
Strict ($+1$)  & $\geq$3 & $\geq$4 & $\geq$2 \& $\geq$3 & $\leq$2 \\
\bottomrule
\end{tabular}
\caption{Threshold configurations for the sensitivity sweep on Synthetic. Each component is shifted by one step; under Lenient the single-implicit-positive clause subsumes the hybrid clause.}
\label{tab:threshold_grid}
\end{table}

Table~\ref{tab:threshold_igf1} reports median IG$_{F1}$ under each setting. As expected, absolute scores rise under Lenient and fall under Strict, but the \emph{relative ordering of models is preserved}. Table~\ref{tab:threshold_rho} quantifies this with the Spearman rank correlation against the Base ranking: all four metrics retain $\rho \geq 0.81$ under both perturbations, and the IG$_R$ and IS rankings are perfectly preserved Base$\to$Strict ($\rho = 1.0$). The Base column reproduces the main-paper values within 0.3pp.

\begin{table}[t]
\centering
\small
\setlength{\tabcolsep}{5pt}
\begin{tabular}{l|ccc}
\toprule
\textbf{Model} & \textbf{Lenient} & \textbf{Base} & \textbf{Strict} \\
\midrule
GPT-OSS-120B          & 76.1 & 67.8 & 43.1 \\
DeepSeek-R1           & 60.2 & 56.5 & 39.6 \\
Qwen3-VL-235B         & 75.1 & 50.8 & 34.6 \\
Qwen3-32B             & 54.3 & 46.3 & 33.1 \\
Qwen2.5-72B           & 56.9 & 45.4 & 36.3 \\
\makecell[l]{DeepSeek-R1-\\Distill-70B} & 58.7 & 37.1 & 26.1 \\
Qwen2-7B              & 34.9 & 14.3 & 11.0 \\
\bottomrule
\end{tabular}
\caption{Median IG$_{F1}$ (\%) on Synthetic under the three threshold settings (Table~\ref{tab:threshold_grid}). Absolute scores shift with the thresholds, but the model ordering is stable.}
\label{tab:threshold_igf1}
\end{table}

\begin{table}[t]
\centering
\small
\setlength{\tabcolsep}{6pt}
\begin{tabular}{l|cc}
\toprule
\textbf{Metric} & \textbf{Base vs Lenient} & \textbf{Base vs Strict} \\
\midrule
IG$_P$    & $+0.815$ & $+0.893$ \\
IG$_R$    & $+0.857$ & $+1.000$ \\
IG$_{F1}$ & $+0.821$ & $+0.893$ \\
IS        & $+0.821$ & $+1.000$ \\
\bottomrule
\end{tabular}
\caption{Spearman rank correlation ($\rho$, $n=7$ models) of the per-metric model ranking against the Base configuration. All $\rho \geq 0.81$; IG$_R$ and IS rankings are perfectly preserved under Strict.}
\label{tab:threshold_rho}
\end{table}

\section{Model Details}
\label{app:model_ids}

Table~\ref{tab:model_ids} lists the HuggingFace identifiers for all models used in this work. We evaluate eight open-weight LLMs spanning four model families (DeepSeek, Qwen, Llama, GPT-OSS) and parameter counts from 7B to 235B. All models were served in full bf16 precision on NVIDIA H100 GPUs with 8-way tensor parallelism. A separate model, Llama-3.3-70B-Instruct, serves as the judge for both IG evidence filtering (Section~\ref{sec:ig}) and IS retrieval evaluation (Section~\ref{sec:is}); it is architecturally distinct from the evaluated Llama-3.1-8B-Instruct to avoid circular evaluation.

\begin{table}[t]
\centering
\footnotesize
\setlength{\tabcolsep}{3pt}
\newcommand{\modelid}[1]{\texttt{\raggedright\hyphenchar\font=`\/ #1}}
\begin{tabular}{@{}l@{\hspace{6pt}}p{5cm}@{}}
\toprule
\textbf{Paper Name} & \textbf{HuggingFace Model ID} \\
\midrule
DeepSeek-R1 & \modelid{deepseek-ai/DeepSeek-R1} \\
GPT-OSS-120B & \modelid{openai/gpt-oss-120b} \\
Qwen3-32B & \modelid{Qwen/Qwen3-32B} \\
Qwen3-VL-235B & \modelid{Qwen/Qwen3-VL-235B-A22B-Instruct} \\
R1-Distill-70B & \modelid{deepseek-ai/DeepSeek-R1-Distill-Llama-70B} \\
Qwen2.5-72B & \modelid{Qwen/Qwen2.5-72B-Instruct} \\
Qwen2-7B & \modelid{Qwen/Qwen2-7B-Instruct} \\
Llama-3.1-8B-Instruct & \modelid{meta-llama/Llama-3.1-8B-Instruct} \\
\midrule
Llama-3.3-70B\textsuperscript{\dag} & \modelid{meta-llama/Llama-3.3-70B-Instruct} \\
\bottomrule
\end{tabular}
\caption{HuggingFace model identifiers for the eight evaluated models, followed by the judge model. \textsuperscript{\dag}Judge model for both IG evidence filtering and IS retrieval evaluation.}
\label{tab:model_ids}
\end{table}

\section{Compute Cost Analysis}
\label{app:compute_cost}

This section quantifies the cost of running the GISTBench evaluation pipeline (Section~\ref{sec:pipeline}). We report cost primarily in \emph{tokens}, which are hardware-agnostic and transfer directly to any serving stack or hosted inference API, and then translate that token volume into the GPU resources we expended to produce the benchmark and this paper. We separate the pipeline into two stages: \textbf{Stage~A}, the candidate-LLM inference that generates interest predictions from each user's interaction history; and \textbf{Stage~B}, the benchmark computation that turns those predictions into IG and IS scores via the Llama-3.3-70B-Instruct judge (evidence filtering, retrieval, and taxonomy mapping; Sections~\ref{sec:ig}--\ref{sec:normalization}). Stage~B is the benchmark-intrinsic cost, independent of which candidate an adopter evaluates.

\paragraph{Methodology.}
Token counts are computed with the Llama-3 tokenizer over the exact prompts and generations stored by the pipeline, including the system prompt and every dialog turn. All GPU figures are NVIDIA~H100~SXM5 inference hours.

\subsection{Token Cost}
\label{app:compute_candidate}

\paragraph{Stage~A: candidate interest generation.}
Table~\ref{tab:tokens_stage_a} reports the token volume for the reference candidate, Llama-3.1-8B-Instruct. Generating interests over the full benchmark consumes $\sim$$302$M input and $\sim$$261$M output tokens, with the Synthetic split accounting for most of it ($175$M input, $198$M output): each synthetic user carries a much richer interaction history, which is sliced into many $100$-engagement prompt windows.

\begin{table}[t]
\centering
\small
\setlength{\tabcolsep}{5pt}
\begin{tabular}{l|r|r|r}
\toprule
\textbf{Dataset} & \textbf{Prompts} & \textbf{Input tok.} & \textbf{Output tok.} \\
\midrule
Synthetic    & $46{,}961$ & $175.3$M & $198.3$M \\
KuaiRec      & $14{,}844$ & $78.0$M  & $49.5$M  \\
MIND         & $2{,}062$  & $13.6$M  & $4.6$M   \\
Amazon Music & $126$      & $0.6$M   & $0.1$M   \\
Goodreads    & $2{,}219$  & $34.7$M  & $8.1$M   \\
\midrule
\textbf{Total} & $\mathbf{66{,}212}$ & $\mathbf{302.2}$M & $\mathbf{260.6}$M \\
\bottomrule
\end{tabular}
\caption{Stage~A candidate-inference token cost for the reference candidate (Llama-3.1-8B-Instruct), by dataset. Other candidates' token volume scales with the number of interests they generate (Table~\ref{tab:compute_judge_per_candidate}).}
\label{tab:tokens_stage_a}
\end{table}

\paragraph{Stage~B: benchmark computation.}
Table~\ref{tab:tokens_stage_b} reports the judge token volume, summed across all $8$ evaluated candidates' inputs. Stage~B consumes \textbf{11.1B input} and $0.12$B output tokens, on the order of $35\times$ the Stage~A input volume, because each candidate-predicted interest triggers up to $\sim$$3.3$ judge calls and the IS distractor-pool prompts each carry a $1{,}000$-object pool. Output is tiny in every pass: the judges emit terse verdicts (a relevance bit, or a short list of object ids).

\begin{table}[t]
\centering
\small
\setlength{\tabcolsep}{4pt}
\begin{tabular}{l|r|r|r}
\toprule
\textbf{Judge pass} & \textbf{Prompts} & \textbf{Input tok.} & \textbf{Output tok.} \\
\midrule
IG evidence relevance (Sec.~\ref{sec:ig})       & $2{,}497{,}314$ & $3.27$B & $66.5$M \\
IS distractor filtering (Sec.~\ref{sec:is})      & $38{,}977$      & $2.70$B & $12.2$M \\
IS retrieval (Sec.~\ref{sec:is})                 & $2{,}399{,}632$ & $4.65$B & $34.2$M \\
Taxonomy mapping (Sec.~\ref{sec:normalization})  & $156{,}083$     & $0.39$B & $2.2$M  \\
\midrule
\textbf{Total} & $\mathbf{5{,}092{,}006}$ & $\mathbf{11.07}$B & $\mathbf{117.7}$M \\
\bottomrule
\end{tabular}
\caption{Stage~B benchmark-computation token cost (Llama-3.3-70B-Instruct judge), summed across all $8$ evaluated candidates' inputs. Per-pass subtotals are rounded sampled estimates; the Total is computed over all per-(model, dataset, judge) cells and may not equal the rounded subtotal sum.}
\label{tab:tokens_stage_b}
\end{table}

\subsection{GPU Resource Footprint}
\label{app:compute_benchmark}

We translate the token volume above into the GPU hours we expended on NVIDIA H100~SXM5 GPUs. Stage~A generation for the reference candidate (Llama-3.1-8B-Instruct) consumed $54.3$~H100-hours on $8$~H100 GPUs ($\sim$$1{,}300$ prompts/hour per H100); other candidates scale with model size and serving configuration. The Stage~B benchmark computation comprises the four judge passes of Table~\ref{tab:tokens_stage_b}, executed on $64$~H100 GPUs; Table~\ref{tab:compute_judge_per_pass} reports its per-pass H100-hour cost.

\begin{table}[t]
\centering
\small
\setlength{\tabcolsep}{4pt}
\begin{tabular}{l|c}
\toprule
\textbf{Judge pass} & \textbf{H100-hours} \\
\midrule
IG evidence relevance filtering (Section~\ref{sec:ig})         & $\sim 380$ \\
IS distractor-pool overlap filtering (Section~\ref{sec:is})    & $\sim 115$ \\
IS retrieval (Section~\ref{sec:is})                            & $\sim 245$ \\
Interest-to-taxonomy mapping (Section~\ref{sec:normalization}) & $\sim 20$ \\
\midrule
\textbf{Total (across all $8$ evaluated candidates)} & $\mathbf{\sim 760}$ \\
\bottomrule
\end{tabular}
\caption{Compute cost of the four Llama-3.3-70B-Instruct judge inference passes that constitute the benchmark computation, summed across the inputs from all $8$ candidates evaluated in this paper, executed on $64$ H100 GPUs. The per-candidate decomposition of this $\sim 760$ H100-hour total is given in Table~\ref{tab:compute_judge_per_candidate}.}
\label{tab:compute_judge_per_pass}
\end{table}

\paragraph{Per-candidate cost.}
The benchmark computation cost attributable to a given candidate LLM scales with the candidate's output volume, primarily the number of interests it generates. The number of Llama-3.3-70B judge calls per candidate-predicted interest has a structural upper bound of $\sim3.33$. Two of the four judge passes are tied one-to-one to the interest count by per-interest grouping, namely IG evidence filtering (Section~\ref{sec:ig}) and IS retrieval (Section~\ref{sec:is}), contributing exactly $1$ call each per interest. The remaining two passes contribute amortized fractional calls per interest bounded by their prompt-batching structure: the taxonomy mapping pass batches $3$ globally-unique interest strings per prompt, contributing at most $1/3$ call per interest (the bound is reached only when no interest string recurs across users or candidates); the IS distractor-pool overlap filter bundles all of a single user's interests into one shortlisting prompt per (user, dataset) tuple, contributing at most $1$ call per interest (the bound is reached only when a user produces a single interest in a given dataset).

Table~\ref{tab:compute_judge_per_candidate} reports the per-candidate breakdown for the candidates evaluated in this paper. The per-candidate cost is \emph{not} monotonic in candidate parameter count. It depends on the volume of interest predictions the candidate produces. DeepSeek-R1 incurs the lowest benchmark computation cost in the table because it produces fewer, more concentrated predictions, while the $7$B Qwen2-7B incurs higher cost than far larger models such as DeepSeek-R1 and Qwen3-32B.

\begin{table}[t]
\centering
\small
\setlength{\tabcolsep}{4pt}
\begin{tabular}{l|r|r}
\toprule
\textbf{Candidate} & \textbf{Generated interests} & \textbf{Benchmark computation H100-hr} \\
\midrule
DeepSeek-R1                       & $35{,}550$ & $\sim 56$ \\
Qwen3-32B                         & $37{,}960$ & $\sim 60$ \\
DeepSeek-R1-Distill-70B     & $48{,}040$ & $\sim 76$ \\
Qwen2-7B                          & $61{,}400$ & $\sim 97$ \\
Qwen3-VL-235B                     & $65{,}290$ & $\sim 103$ \\
Llama-3.1-8B-Instruct             & $68{,}496$ & $\sim 108$ \\
Qwen2.5-72B                       & $78{,}358$ & $\sim 123$ \\
GPT-OSS-120B                      & $86{,}848$ & $\sim 137$ \\
\bottomrule
\end{tabular}
\caption{Benchmark computation cost per evaluated candidate, summing to the $\sim 760$ H100-hour aggregate in Table~\ref{tab:compute_judge_per_pass}. Cost scales with each candidate's generated interest count.}
\label{tab:compute_judge_per_candidate}
\end{table}

\subsection{Total GPU Footprint}
\label{app:compute_total}

The benchmark-intrinsic GPU cost, paid once per evaluation independent of the candidate, is the $\sim$$760$ H100-hours of Stage~B judge computation (Table~\ref{tab:compute_judge_per_pass}), executed on $64$~H100 GPUs. Evaluating a single candidate end-to-end costs that model's Stage~A generation (Section~\ref{app:compute_candidate}) plus its per-candidate Stage~B share (Table~\ref{tab:compute_judge_per_candidate}); for the reference Llama-3.1-8B-Instruct this is $54 + 108 = 162$ H100-hours. The total GPU resource expended to produce every result in this paper is therefore the $\sim$$760$ H100-hour judge computation plus the eight candidates' Stage~A generation runs, a footprint dominated by the Stage~B judge computation, which is the reusable, candidate-independent core of the benchmark.

\paragraph{USD pricing reference.}
The analysis above quantifies the H100-hours required; the corresponding dollar cost depends on the GPU pricing available to the adopter. As a rough reference, as of May~2026, specialist GPU-cloud providers such as Together~AI and Lambda~Labs tend to offer H100s at approximately \$$2$--\$$3$ per H100-hour on spot or reserved tiers; on-demand H100 pricing from hyperscalers such as AWS, GCP, and Azure is typically somewhat higher. Of course, this pricing varies across providers, regions, and reservation tiers. These figures are rough estimates drawn from publicly available pricing pages and should be verified with each provider for current rates.

For the candidate-LLM interest generation cost (Section~\ref{app:compute_candidate}), token usage scales with the size of the dataset on which the candidate is asked to generate interests; adopters who use a hosted-inference API for the candidate model rather than self-host on H100s can estimate this cost directly from the per-token pricing offered by the relevant API provider.

\paragraph{Frontier candidates (API-metered).} The three proprietary frontier candidates (GPT-5, Claude 4.6, and Gemini 3.5 Flash) are not self-hosted on H100s but accessed through hosted APIs billed per token, so they fall outside the H100-hour accounting above; the tables in this appendix cover the eight open-weight candidates only. Their Stage~A candidate-generation cost is estimated on the same per-token basis as any hosted candidate: token volume follows the per-user prompt profile of Table~\ref{tab:tokens_stage_a} (one or more $100$-engagement prompt windows per user), so the dollar cost is that volume multiplied by the provider's per-token input and output prices. Because those prices differ by provider, model tier, and token direction, and change frequently, we do not fix a dollar figure here; adopters should read the current price sheet of the chosen API. The benchmark-intrinsic Stage~B judge cost (Table~\ref{tab:compute_judge_per_pass}) is unaffected by this choice: it runs on our own H100s over each candidate's predicted interests regardless of whether the candidate was open-weight or API-served.

\subsection{Cheaper Evaluation via User Subsampling}
\label{app:cheaper_eval}

\begin{figure}[t]
\centering
\includegraphics[width=0.6\columnwidth]{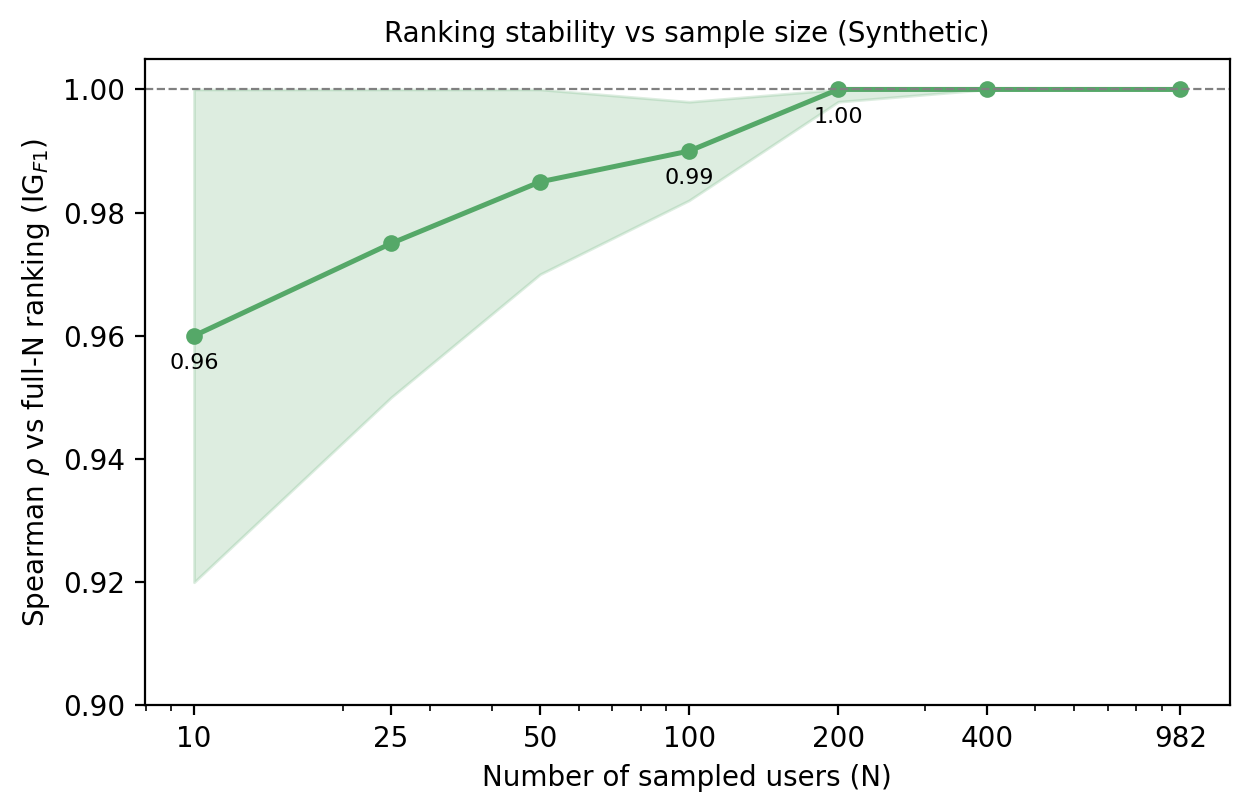}
\caption{Ranking stability versus sample size on the Synthetic split. For each $N$ we draw $50$ random subsamples of $N$ users, rank the models by median IG$_{F1}$ within each, and compute the Spearman correlation against the full-set ($N=982$) ranking; the line is the mean and the shaded band is $\pm 1$ standard deviation across replicates. The ranking is recovered to $\rho=0.96$ at $N=10$ and to $1.00$ by $N=200$.}
\label{fig:pareto_ranking}
\end{figure}

On the Synthetic split, the GISTBench ranking saturates with far fewer users than the full set, so adopters can compare candidate models at a fraction of the cost. We subsample $N$ users uniformly (50 replicates each) and recompute all metrics with the IG Recall oracle re-derived per subset. The metric \emph{means} are essentially flat from $N=10$ onward; only their replicate-to-replicate uncertainty contracts, at roughly the $1/\sqrt{N}$ rate (the IG$_{F1}$ standard deviation falls from $2.9$ percentage points at $N=10$ to $0.7$ at $N=100$ and ${\sim}0$ at $N=982$). Ranking agreement with the full-set ($N=982$) ordering, measured by Spearman $\rho$ over IG$_{F1}$ (Figure~\ref{fig:pareto_ranking}), is already $\rho=0.96$ at $N=10$, $0.99$ at $N=100$, and reaches $1.00$ from $N=200$. Two regimes follow. Adopters who need only a \emph{ranking} of candidate models can operate at $N\approx25$--$100$ users, roughly $1.5\%$ of the full Stage~A token budget (for $\rho\geq0.96$) up to about $10\%$ (for $\rho\geq0.99$). Adopters who need \emph{absolute} metric values reach sub-percentage-point standard error by $N\approx100$--$200$ users, about $5\times$ cheaper than the full evaluation.

\end{document}